\documentclass{article}

\PassOptionsToPackage{numbers, compress}{natbib}

\usepackage[preprint]{neurips_2025}

\usepackage[utf8]{inputenc} 
\usepackage[T1]{fontenc}    
\usepackage{hyperref}       
\usepackage{url}            
\usepackage{booktabs}       
\usepackage{amsfonts}       
\usepackage{nicefrac}       
\usepackage{microtype}      
\usepackage{xcolor}         

\usepackage{multirow}
\usepackage{subcaption}
\usepackage{microtype}
\usepackage{ragged2e}
\usepackage[most]{tcolorbox}

\newtcolorbox{promptbox}[2][]{%
  enhanced,
  breakable,
  width=\textwidth,    
  colback=gray!6,
  boxrule=1.5pt,
  left=6pt,right=6pt,top=6pt,bottom=6pt,
  fonttitle=\bfseries,
  title={#2},
  before skip=6pt, after skip=6pt,
  #1
}

\newtcolorbox{snippetbox}{%
  enhanced, breakable,
  colback=gray!10, colframe=gray!30,
  boxrule=0pt, leftrule=3pt, arc=1pt,
  left=4pt,right=4pt,top=3pt,bottom=3pt,
  fontupper=\small,
  before upper=\RaggedRight\setlength{\parindent}{0pt},
}

\newtcolorbox{blueblock}{%
  enhanced,
  breakable,
  colback=blue!8,
  colframe=blue!35,
  boxrule=0.6pt,
  arc=1pt,
  left=6pt,right=6pt,top=6pt,bottom=6pt,
  before skip=6pt, after skip=6pt,
}

\newtcolorbox{grayblock}{%
  enhanced,
  breakable,
  colback=gray!14,      
  colframe=gray!45,
  boxrule=0.6pt,
  arc=1pt,
  left=6pt,right=6pt,top=6pt,bottom=6pt,
  before skip=6pt, after skip=6pt,
}

\usepackage{amsthm}
\newtheorem{definition}{Definition}
\usepackage[inline]{enumitem}
\usepackage{booktabs}       
\usepackage{amssymb}

\title{Hallucination-Resistant Relation Extraction via Dependency-Aware Sentence Simplification and Two-tiered Hierarchical Refinement}

\author{%
  \textbf{Yupei Yang\textsuperscript{1,4}\thanks{This work was done when the author was a research intern at Alibaba Group.}},~
  \textbf{Fan Feng\textsuperscript{2,3}},~
  \textbf{Lin Yang\textsuperscript{4}},~
  \textbf{Wanxi Deng\textsuperscript{4}},~
  \textbf{Lin Qu\textsuperscript{4}}\\
  \textbf{Biwei Huang\textsuperscript{2}},~
  \textbf{Shikui Tu\textsuperscript{1}\thanks{Corresponding author}},~
  \textbf{Lei Xu\textsuperscript{1}} \\[0.25cm]
  \textsuperscript{1}Shanghai Jiao Tong University \qquad
  \textsuperscript{2}University of California San Diego \\
  \textsuperscript{3}Mohamed bin Zayed University of Artificial Intelligence \qquad
  \textsuperscript{4}Alibaba Group \\[0.15cm]
  \texttt{\{yupei\_yang,tushikui,leixu\}@sjtu.edu.cn, bih007@ucsd.edu}\\
  \texttt{ffeng1017@gmail.com, zhihe.yyp@alibaba-inc.com}
}

\begin{document}

\maketitle

\begin{abstract}
  Relation extraction (RE) enables the construction of structured knowledge for many downstream applications. While large language models (LLMs) have shown great promise in this task, they often struggle to reliably determine whether a relation exists, particularly in sentences with complex syntax or subtle semantics. For instance, we find that \texttt{Qwen2.5-14B-Instruct} incorrectly predicts a relation in 96.9\% of \texttt{NO-RELATION} instances on SciERC, revealing a severe hallucination problem. To address these challenges, we propose \textbf{DEPTH}, a framework that integrates \textit{Dependency-aware sEntence simPlification} and \textit{Two-tiered Hierarchical refinement} into the relation extraction pipeline. Given a sentence and its candidate entity pairs, DEPTH operates in two stages:
  \begin{enumerate*}[label=(\arabic*)]
      \item the \textit{Grounding} module extracts relations for each pair by leveraging their shortest dependency path, distilling the sentence into a minimal yet coherent relational context that reduces syntactic noise while preserving key semantics;
      \item the \textit{Refinement} module aggregates all local predictions and revises them based on a holistic understanding of the sentence, correcting omissions and inconsistencies.
  \end{enumerate*}
  We further introduce a causality-driven reward model that mitigates reward hacking by disentangling spurious correlations, enabling robust fine-tuning via reinforcement learning with human feedback. Experiments on eight well-established benchmarks demonstrate that DEPTH reduces the average hallucination rate to 7.9\% while achieving a 9.3\% improvement in average F1 score over existing LLM-based extraction baselines.
\end{abstract}

\section{Introduction}
Relation extraction, the task of identifying semantic relationships between entities in unstructured text, serves as a fundamental component in natural language processing (NLP) \cite{bach2007review,pawar2017relation,detroja2023survey,zhao2024comprehensive}. Its applications span a wide range of domains, including social media analysis \cite{zheng2021mnre}, knowledge graph construction \cite{yu2020relationship}, and question answering \cite{xu2016question}. Traditional relation extraction methods typically rely on large-scale annotated datasets, making them costly to deploy and often limited in generalization. Recently, LLMs, empowered by increased model capacity and massive pretraining corpora, have demonstrated impressive capabilities in text understanding and generation. As a result, leveraging LLMs for relation extraction has emerged as a promising research direction \cite{zhang2023aligning,li2023revisiting,li2024llm,zhou2024leap}.

\begin{table}[t]
  \centering
  \resizebox{\linewidth}{!}{
  \begin{tabular}{c*{16}{c}}
    \toprule
    \multirow{3}{*}{Metric}
      & \multicolumn{2}{c}{\texttt{USED-FOR}}
      & \multicolumn{2}{c}{\texttt{FEATURE-OF}}
      & \multicolumn{2}{c}{\texttt{HYPONYM-OF}}
      & \multicolumn{2}{c}{\texttt{PART-OF}}
      & \multicolumn{2}{c}{\texttt{EVALUATE-FOR}}
      & \multicolumn{2}{c}{\texttt{COMPARE}}
      & \multicolumn{2}{c}{\texttt{CONJUNCTION}}
      & \multicolumn{2}{c}{\texttt{NO-RELATION}}\\
    \cmidrule(l){2-3}\cmidrule(l){4-5}\cmidrule(l){6-7}\cmidrule(l){8-9}
    \cmidrule(l){10-11}\cmidrule(l){12-13}\cmidrule(l){14-15}\cmidrule(l){16-17}
      & \quad$\mathcal{D}_1$ & $\mathcal{D}_1^+$
      & \quad$\mathcal{D}_1$ & $\mathcal{D}_1^+$
      & \quad$\mathcal{D}_1$ & $\mathcal{D}_1^+$
      & \quad$\mathcal{D}_1$ & $\mathcal{D}_1^+$
      & \quad$\mathcal{D}_1$ & $\mathcal{D}_1^+$
      & \quad$\mathcal{D}_1$ & $\mathcal{D}_1^+$
      & \quad$\mathcal{D}_1$ & $\mathcal{D}_1^+$
      & \quad$\mathcal{D}_1$ & $\mathcal{D}_2$\\
    \midrule
    TP & \quad167 & 186 & \quad21 & 23 & \quad3 & 2 & \quad30 & 31 & \quad82 & 81 & \quad32 & 33 & \quad26 & 21 & \quad-- & 45 \\
    FP & \quad17 & \textbf{198} & \quad153 & \textbf{427} & \quad1 & \textbf{3} & \quad103 & \textbf{263} & \quad248 & \textbf{851} & \quad54 & \textbf{199} & \quad0 & \textbf{14} & \quad-- & 0 \\
    FN & \quad376 & 357 & \quad40 & 38 & \quad64 & 65 & \quad33 & 32 & \quad10 & 11 & \quad2 & 1 & \quad94 & 99 & \quad-- & \textbf{1430} \\
    \bottomrule
  \end{tabular}}
  \caption{True positives (TP), false positives (FP), and false negatives (FN) by relation category on SciERC.}
  \label{tab:relation-metrics}
\end{table}

However, a critical limitation of LLM-based approaches is their tendency to overpredict relations: even when no meaningful connection exists between an entity pair, LLMs often hallucinate a spurious relation rather than outputting \texttt{NO-RELATION}. For example, when evaluated on the SciERC dataset, \texttt{Qwen2.5-14B-Instruct} correctly identifies \texttt{NO-RELATION} in only 45 out of 1,475 instances, misclassifying the remaining 1,430 as having some relation (see Table \ref{tab:relation-metrics}). These false positives (FPs) introduce substantial noise into extracted knowledge and can severely compromise the quality and reliability of downstream applications, especially when applied at enterprise-scale document processing. Unfortunately, despite strong performance on standard relation extraction benchmarks, most existing methods rarely emphasize or rigorously evaluate this ability to say ``no''.

To this end, we propose \textbf{DEPTH}, a framework that integrates \textit{Dependency-aware sEntence simPlification} and \textit{Two-tiered Hierarchical refinement} into the extraction pipeline to reduce hallucinated relations and improve extraction fidelity. Specifically, DEPTH decomposes the task into two stages: \textit{Grounding} and \textit{Refinement}. The \textit{Grounding} module focuses on local extraction, aiming to improve prediction accuracy for each candidate entity pair. Motivated by the observation that the essential information needed to infer a relation between two entities is often captured by the shortest dependency path (SDP) connecting them in the dependency tree \cite{culotta2004dependency,bunescu2005shortest,fundel2007relex,xu2015semantic,xue5131704multi}, we perform dependency parsing \cite{dozat2016deep,wang2018neural,li2018seq2seq} on the input sentence and simplify it by retaining only the SDP and its immediate context. This reduces distracting information while preserving the relational semantics. In addition, we explicitly encode dependency-based cues as natural language prompts to guide LLM during relation prediction.

While the \textit{Grounding} module effectively reduces syntactic noise through sentence simplification and dependency-guided prompting, we find that relying solely on localized context remains insufficient. LLMs still tend to infer relations based on superficial lexical patterns, such as the frequent co-occurrence of certain entity pairs. This reliance on shallow statistical cues, rather than genuine semantic reasoning, often leads to systematic hallucinations. To mitigate this issue, a promising approach is reinforcement learning with human feedback (RLHF) \cite{ouyang2022training,sun2023aligning}. However, the success of RLHF crucially depends on the robustness of the reward model (RM). Existing RM training methods, unfortunately, are often vulnerable to reward hacking \cite{amodei2016concrete,gao2023scaling,skalse2022defining,eisenstein2023helping,wang2025beyond}, whereby the model captures spurious correlations instead of truly causal signals. To address this, we adopt a causal perspective on the reward modeling process for relation extraction. Specifically, we construct a causal graph of the reward model and introduce a causal factorization method that separates each prompt-response pair into reward-relevant and reward-irrelevant components. By training the reward model solely on the relevant parts, we encourage it to focus on features truly indicative of relation existence, while discarding confounding patterns that do not causally affect the reward. We then apply Proximal Policy Optimization (PPO) \cite{schulman2017proximal,stiennon2020learning,bai2022training} to fine-tune the LLM based on this robust reward model, significantly enhancing the reliability of relation predictions.

On the other hand, the \textit{Refinement} module operates at the sentence level by aggregating the predictions generated by the \textit{Grounding} module across all candidate entity pairs. We then prompt the LLM to reassess these predictions under global semantic constraints (such as relational transitivity or mutual exclusivity), to identify and correct missing, erroneous, or logically inconsistent relations. By incorporating a global view of the sentence, the \textit{Refinement} module equips the model with a form of self-correction, allowing it to reconcile local predictions with broader sentence-level coherence, which we find to be highly effective in mitigating hallucinated relations in our experiments. To summarize, our main contributions are as follows:
\begin{itemize}[leftmargin=12pt,topsep=0pt,itemsep=0pt]
    \item We propose \textbf{DEPTH}, a two-tiered hierarchical framework that decouples entity-pair grounding from sentence-level refinement: the \textit{Grounding} module performs dependency-aware sentence simplification based on SDPs to distill a minimal yet coherent relational context for each entity pair, while the \textit{Refinement} module aggregates all local predictions and revises them under sentence-level semantic constraints to recover omissions and resolve inconsistencies.
    \item We tailor dependency parsing to LLM-based RE via a dependency-aware simplification procedure: for each entity pair, we
    \begin{enumerate*}[label=(\arabic*)]
        \item extract the SDP using a standard dependency parser,
        \item distill the sentence into an SDP-centered context, 
        \item convert the SDP into a natural-language description as structural cues to guide LLM prediction, thereby reducing hallucinations.
    \end{enumerate*}
    \item We develop a causality-driven reward modeling framework that mitigates reward hacking by explicitly disentangling reward-relevant signals from reward-irrelevant factors through a fixed, template-based factorization, yielding a robust reward model that resists spurious correlations and enables more effective policy optimization.
    \item Extensive experiments on eight benchmarks demonstrate that DEPTH reduces the average hallucination rate to 7.9\% and improves average micro-F1 by 9.3\% over existing LLM-based baselines. These results highlight DEPTH’s superior capability in reliably distinguishing relation existence, a critical requirement for real-world knowledge extraction where false positives are particularly costly.
\end{itemize}

\section{Preliminaries and Problem Setup}\label{sec:set_up}
Given a sentence $s$ containing an entity pair $(e_1, e_2)$, relation extraction aims to identify the semantic relation $\rho \in \mathcal{R}$ that holds between them, where $\mathcal{R}$ denotes a predefined set of relation types. In the LLM-based setting, the sentence $s$, entity pair $(e_1,e_2)$, and relation set $\mathcal{R}$ are encoded into a prompt $x$ for an LLM. The LLM then generates a textual response $y$, from which the predicted relation $\hat{\rho}\in\mathcal{R}$ is extracted.

\begin{table}[t]
  \centering
  \begin{tabular}{lcccc}
    \toprule
    Split & P $\uparrow$ & R $\uparrow$ & F1 $\uparrow$ & HR $\downarrow$\\
    \midrule
    $\mathcal{D}_1$ (positive-only) & 38.5 & 36.8 & 37.7 & -- \\
    $\mathcal{D}_1^+$ (positives within full) & 16.2 & 38.5 & 22.8 & --\\
    $\mathcal{D}_2$ (\texttt{NO-RELATION} within full) & 100.0 & 3.1 & 5.9 & \textbf{96.9}\\
    \bottomrule
  \end{tabular}
  \caption{Zero-shot extraction results on SciERC under different evaluation protocols.}
  \label{tab:split-metrics}
\end{table}

An essential, yet often overlooked, requirement in this process is the ability of LLMs to return \texttt{NO-RELATION} when no suitable relation in $\mathcal{R}$ applies.
We study this issue on SciERC \cite{luan2018multi}. Let $\mathcal{D}$ be the full evaluation set containing both positive and negative instances, we define two disjoint subsets: $\mathcal{D}_1=\{(e_1,\rho,e_2)\in\mathcal{D}:\rho\in\mathcal{R}\}$ and $\mathcal{D}_2=\{(e_1,\rho,e_2)\in\mathcal{D}:\rho=\texttt{NO-RELATION}\}$. Specifically, we consider two evaluation protocols:
\begin{itemize}[leftmargin=12pt,topsep=0pt,itemsep=0pt]
    \item Positive-only: evaluate the LLM on $\mathcal{D}_1$ with label set $\mathcal{R}$ (i.e., \texttt{NO-RELATION} is excluded).
    \item Full: evaluate on $\mathcal{D}$ with full label set $\mathcal{R}\cup\{\texttt{NO-RELATION}\}$. For interpretability, we additionally report performance restricted to the $\mathcal{D}_1$ portion (denoted $\mathcal{D}_1^+$) and the $\mathcal{D}_2$ portion within this full evaluation.
\end{itemize}
To quantify false assertions of relations on \texttt{NO-RELATION} instances, we use:
\begin{definition}[Hallucination Rate (HR)]
For the \texttt{NO-RELATION} class, the hallucination rate is
\begin{equation}
  \mathrm{HR} = \frac{\mathrm{FN}}{\mathrm{TP}+\mathrm{FN}},
\end{equation}
where $\mathrm{TP}$ is the number of \texttt{NO-RELATION} instances correctly predicted as \texttt{NO-RELATION}, and $\mathrm{FN}$ is the number of \texttt{NO-RELATION} instances misclassified as having any relation in $\mathcal{R}$. HR is defined exclusively for the \texttt{NO-RELATION} class and equals the complement of its recall: $\mathrm{HR} = 1 - \mathrm{Recall}_{\texttt{NO-RELATION}}$.
\end{definition}

\begin{figure}
  \centering
  \includegraphics[width=\linewidth]{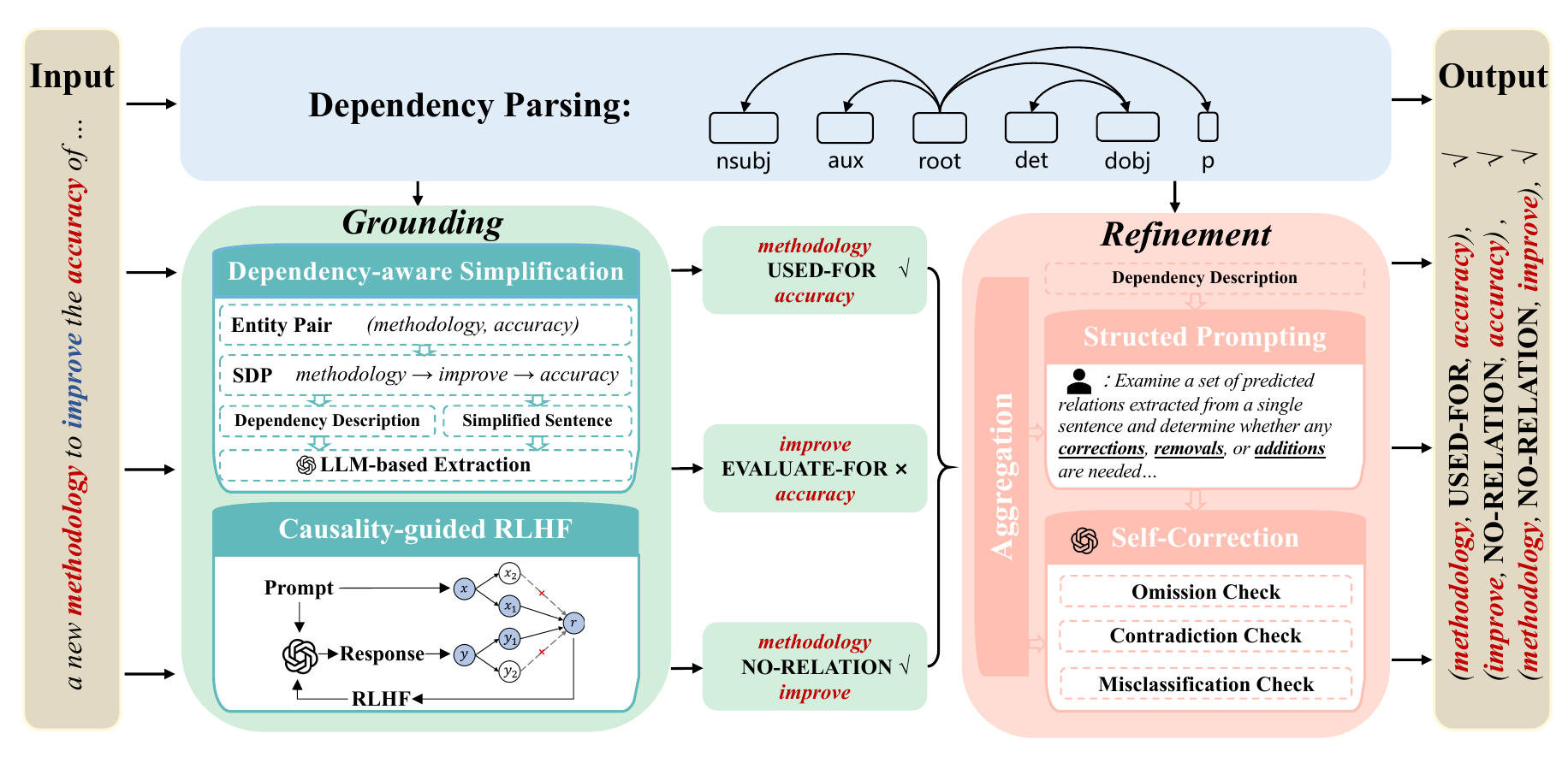}
  \caption{Overall framework of DEPTH: the \textit{Grounding} module uses SDP-based simplification and an RLHF-fine-tuned LLM to predict per-pair relations (only three are illustrated for brevity), while the \textit{Refinement} module jointly refines all predictions using global context. Dependency parsing provides structural guidance throughout, and RLHF is applied only during training.}
  \label{fig:framework}
\end{figure}

Table~\ref{tab:split-metrics} reports zero-shot results of \texttt{Qwen2.5-} \texttt{14B-Instruct}, with detailed TP, FP, and FN breakdowns presented in Table~\ref{tab:relation-metrics}. We observe an extremely high HR of 96.9\% on $\mathcal{D}_2$, which means that most \texttt{NO-RELATION} instances are wrongly assigned some relation. Moreover, comparing $\mathcal{D}_1$ to $\mathcal{D}_1^+$, we observe a substantial increase in false positives with negligible change in true positives. This indicates the model remains reasonably accurate on truly related pairs, but frequently hallucinates relations when confronted with unrelated pairs, thereby underscoring the challenge of reliably detecting relation absence in realistic settings where \texttt{NO-RELATION} is prevalent.

\section{Methodology}
To address the challenges discussed above, we propose DEPTH, a framework that not only classifies semantic relations accurately, but also equips LLMs with the critical ability to discern whether a relation exists. Given a sentence and candidate entity pairs, DEPTH first uses the \textit{Grounding} module to simplify the sentence via SDPs and predict relations with an LLM fine-tuned through causality-guided RLHF. The \textit{Refinement} module then aggregates predictions across all pairs and performs global validation and correction. Dependency parsing supports both modules by providing necessary syntactic structures. The overall workflow of DEPTH is illustrated in Figure \ref{fig:framework}. In the remainder of this section, we elaborate on each component in detail.

\subsection{Dependency-aware Simplification for Entity-Pair Grounding}
Previous works typically focus on LLM-centric approaches, such as prompt engineering and supervised fine-tuning, often overlooking the intrinsic value of syntactic structure in the sentence itself. In other words, existing methods perform relation extraction on raw sentences. However, when sentences are long or particularly complex, LLMs, especially lightweight models, struggle with precise semantic understanding of sentence details. As a result, hallucination issues often arise during relation extraction.

\begin{figure}[t]
  \centering
  \includegraphics[width=.9\linewidth]{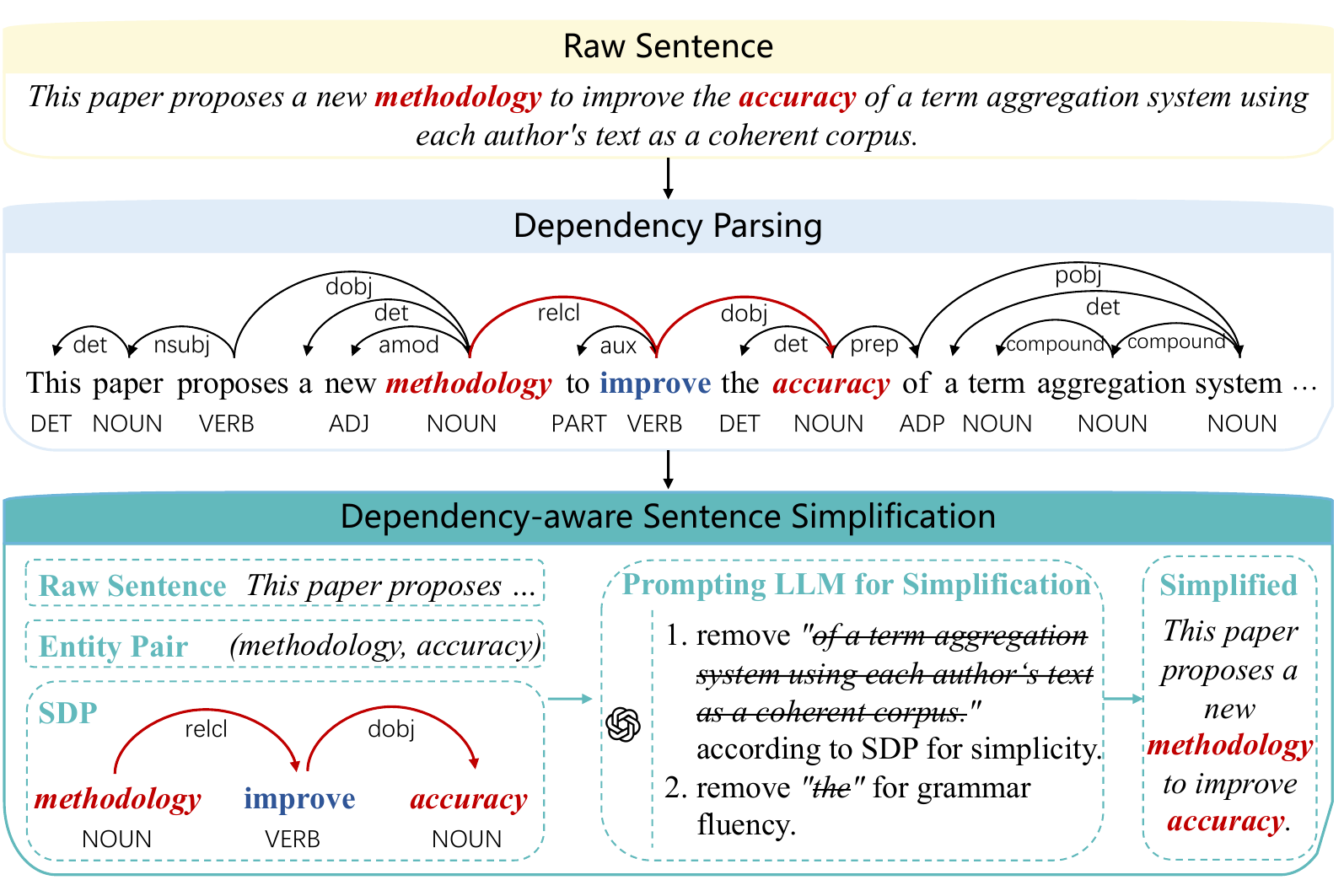}
  \caption{Illustration of the Dependency-aware Simplification module, which leverages the SDP to produce a concise relational context for extraction. The corresponding prompt template is provided in Appendix~\ref{sec:ape_template}.}
  \label{fig:dependency-tree}
\end{figure}

Several studies in deep learning have shown that dependency parsing provides valuable insights, particularly when entities are distant in a sentence, helping the model better capture and explain relations \cite{fundel2007relex,hwang2020spatial}. Building on this, we apply dependency parsing in LLM-based relation extraction to simplify sentence structures and reduce extraction complexity. For a given sentence $s$, we first utilize \texttt{spaCy} \cite{Honnibal_spaCy_Industrial-strength_Natural_2020}, an open-source dependency parsing library, to generate its dependency tree. This tree encodes the syntactic relations between all words in $s$ as directed edges. Figure \ref{fig:dependency-tree} illustrates the dependency tree for the sentence: \textit{``This paper proposes a new methodology to improve the accuracy of a term aggregation system using each author's text as a coherent corpus.''}

For the entity pair \textit{``methodology''} and \textit{``accuracy''}, we extract their SDP: \textit{``methodology → improve → accuracy''}. Since the necessary information to identify the relation between entities is typically contained in the SDP, we then prompt the LLM to simplify the sentence by retaining only the essential parts and removing irrelevant details. As an example, the sentence in Figure \ref{fig:dependency-tree} would be transformed into: \textit{``This paper proposes a new methodology to improve accuracy.''} To ensure that the LLM comprehends the dependency between entities without losing essential information, we further convert the SDP into a textual description, thereby enhancing the model’s understanding. The simplified sentence, along with its corresponding SDP description, is then fed into the LLM for relation extraction.

\subsection{Causal Reward Modeling for Hallucination-Resistant Extraction}
\begin{figure}[t]
  \centering
  \includegraphics[width=\linewidth]{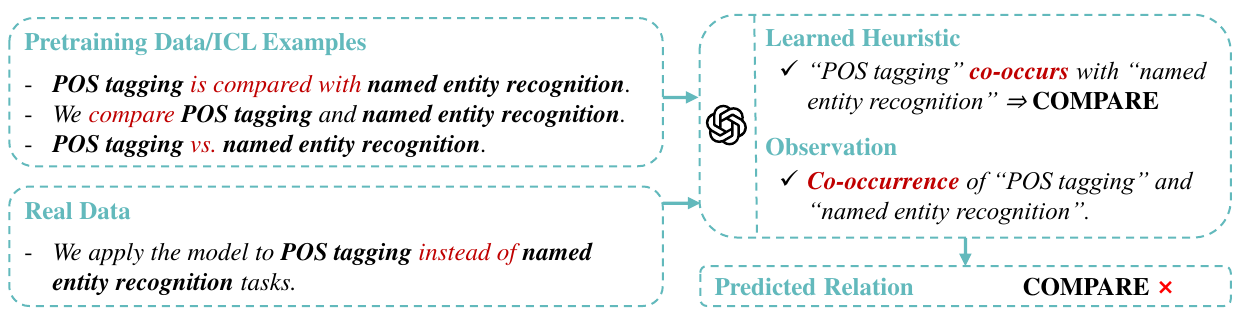}
  \caption{An example of hallucination in LLM-based extraction caused by co-occurrence.}
  \label{fig:Causal-RM}
\end{figure}

Beyond insufficient semantic comprehension, another critical factor leading to biased LLM outputs is their tendency to infer relationships through superficial lexical patterns (e.g., high-frequency co-occurrence of phrases). Once spurious correlations are learned during pre-training or through prompting, the presence of identical phrases in real-world samples inevitably triggers incorrect relational inferences. Figure \ref{fig:Causal-RM} illustrates an in-context learning (ICL) example where the LLM incorrectly infers that the co-occurrence of \textit{``POS tagging''} and \textit{``named entity recognition''} indicates a \texttt{COMPARE} relation. Therefore, given the sentence \textit{``We apply the model to POS tagging instead of named entity recognition tasks''}, the model invariably predicts the relation between \textit{``POS tagging''} and \textit{``named entity recognition''} as \texttt{COMPARE}, even though the true relation should be \texttt{NO-RELATION}.

\begin{figure}[t]
  \centering
  \begin{subfigure}[t]{0.25\linewidth}
    \centering
    \includegraphics[width=\linewidth]{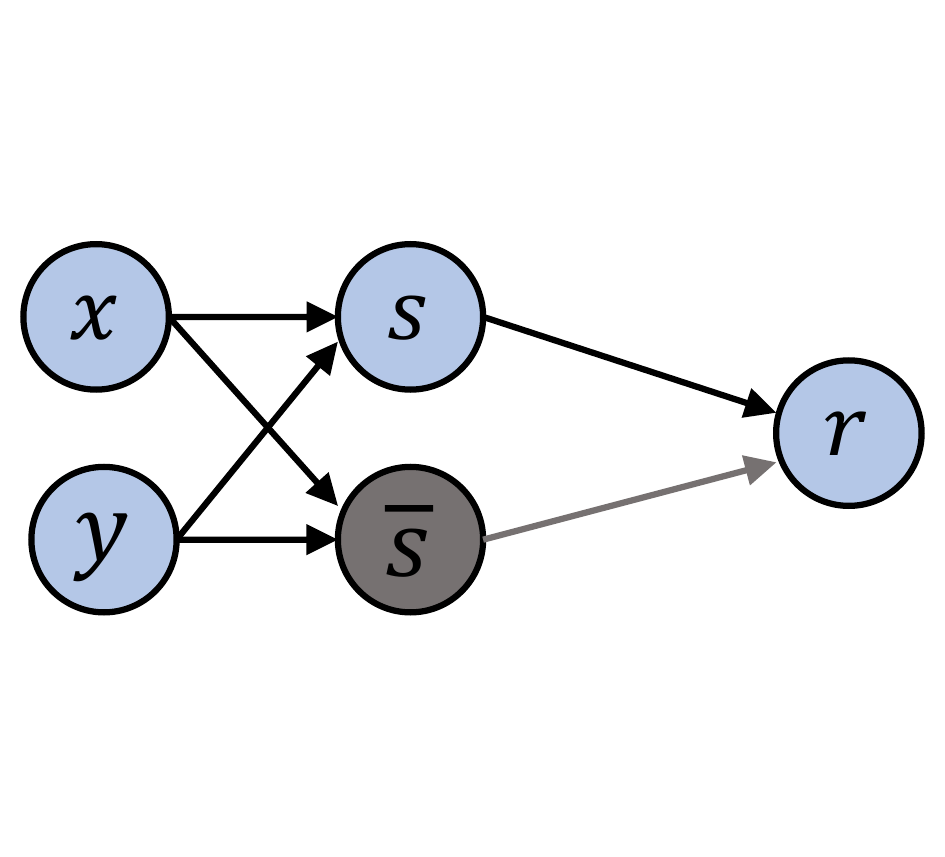}
    \caption{Standard RM.}
    \label{fig:Causal-RM-a}
  \end{subfigure}%
  \hspace{2cm}
  \begin{subfigure}[t]{0.25\linewidth}
    \centering
    \includegraphics[width=\linewidth]{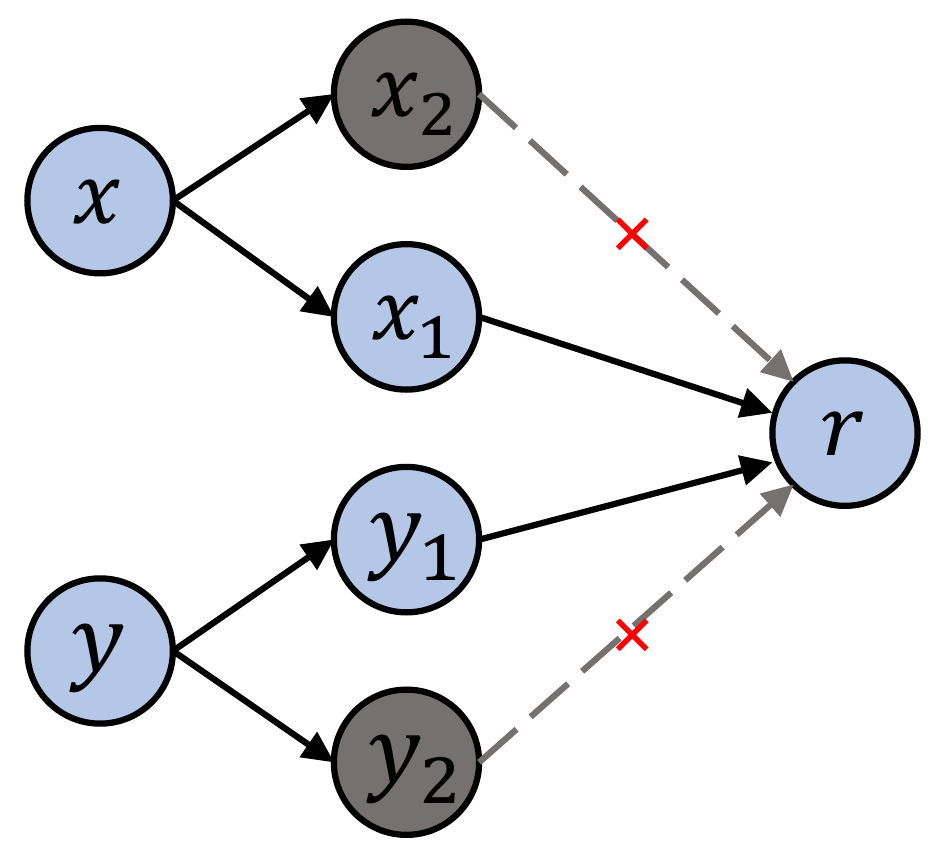}
    \caption{Causal RM.}
    \label{fig:Causal-RM-b}
  \end{subfigure}
  \caption{Causal diagrams of the standard RM training process and our causal method.}
\end{figure}

A widely adopted approach to aligning LLMs with intended behaviors is RLHF, which fine-tunes the policy using rewards predicted by a learned reward model. However, the effectiveness of RLHF heavily depends on the reliability of the reward model, which is often susceptible to reward hacking--where the model assigns high scores based on spurious correlations rather than genuine alignment with the target objectives. To uncover the root cause of this issue, we examine the reward model training process through a causal lens. As shown in Figure \ref{fig:Causal-RM-a}, let $x$ denote the prompt, $y$ the corresponding response, and $r$ the reward. We decompose the input pair $(x, y)$ into two components: $s$, the reward-relevant factors, and $\overline{s}$, the reward-irrelevant ones. Here, $s$ contains the essential information needed to learn an ideal reward function, while $\overline{s}$ captures confounding signals, such as sequence length or stylistic patterns. This formulation highlights that reward hacking arises from the unintended causal path $\overline{s} \to r$. Hence, a robust reward model must explicitly block this influence to avoid learning from spurious correlations.

To achieve this, we decompose the prompt and response as $x = (x_1, x_2)$ and $y = (y_1, y_2)$, where $(x_1, y_1)$ contains reward-relevant information and $(x_2, y_2)$ captures reward-irrelevant factors. Specifically, $x_1$ is defined as the minimal component of the \textit{Grounding} prompt sufficient for reward prediction, consisting of the task definition, simplified sentence, target entity pair, and a natural-language description of their dependency path; while $x_2$ is the remaining prompt content includes in-context examples and auxiliary instructions. For the LLM-generated output, $y_1$ is restricted to the predicted relation label $\hat{\rho}$ in a fixed JSON field, and $y_2$ comprises any additional generated text, such as explanatory rationales. The exact templates for $x_1$ and $y_1$, along with full implementation details, are provided in Appendix~\ref{sec:ape_rm_training}.


During the training process, we optimize the reward model using only $x_1$ and $y_1$, effectively eliminating spurious correlations and transitioning from the causal structure in Figure \ref{fig:Causal-RM-a} to that in Figure \ref{fig:Causal-RM-b}. A key advantage of this approach is that the factorization is applied at the input level and is fixed \emph{a priori} via predefined templates rather than learned from data, thereby requiring no changes to the training architecture and incurring no additional computational cost. Experimental results in Section \ref{sec:exp} validate the quality and robustness of the learned reward model.

\subsection{Global Consistency Refinement via Structured Prompting}
While the \textit{Grounding} module significantly improves relation prediction accuracy for individual entity pairs, treating each pair in isolation can lead to inconsistencies or omissions due to the absence of global semantic constraints. Therefore, we introduce the \textit{Refinement} module, which aggregates local predictions and integrates sentence-level dependency structures to form a hierarchical correction mechanism. This enables global consistency enforcement across predicted relations and facilitates the recovery of overlooked relations.

For a given sentence $s$, the \textit{Refinement} module first aggregates all candidate relations predicted by the \textit{Grounding} module (see Figure \ref{fig:framework}). We then prompt the LLM to convert the dependency tree of $s$, generated by \texttt{spaCy}, into a natural language description using the structured instruction provided in Appendix~\ref{sec:ape_template}. This description, combined with the aggregated entity pairs, forms the input prompt for the LLM, guiding it to perform three-stage semantic calibration: (1) Omission check: identify relations overlooked by the initial predictions by leveraging the full-sentence syntactic structure; (2) Contradiction check: detect and resolve logically inconsistent relations (e.g., mutually exclusive facts), retaining the most coherent prediction based on dependency and context; (3) Misclassification check: correct errors arising from local context bias. The refinement process is performed via ICL using the RLHF‑fine-tuned LLM. Detailed prompt templates are provided in Appendix \ref{sec:ape_template}.

\section{Experiments}\label{sec:exp}
In this section, we present the experimental results of DEPTH on various relation extraction benchmarks. Specifically, our evaluation is designed to answer the following research questions:
\begin{itemize}[leftmargin=12pt,topsep=0pt,itemsep=0pt]
\item \textbf{Q1}: Can DEPTH reduce hallucinations when determining whether a relation exists between two entities?
\item \textbf{Q2}: Does this reduction in hallucinations contribute to improved overall extraction performance?
\item \textbf{Q3}: Is the learned reward model robust to spurious correlations?
\item \textbf{Q4}: Can DEPTH function as a general-purpose relation extraction model that eliminates the need for task-specific training across datasets?
\end{itemize}

\subsection{Experimental Setup}
\paragraph{Datasets.}
We evaluate DEPTH across a diverse collection of relation extraction benchmarks, including four news-domain datasets: TACRED \cite{zhang2017tacred}, TACREV \cite{alt-etal-2020-tacrev}, Re-TACRED \cite{stoica2021re}, and NYT11 \cite{hoffmann2011knowledge}; two scientific-domain datasets: SciERC \cite{luan2018multi} and FOBIE \cite{kruiper2020layman}; one biomedical-domain dataset: DDI \cite{herrero2013ddi}; and one general-domain dataset: SemEval \cite{hendrickx2019semeval}. Detailed descriptions of each dataset are provided in Appendix \ref{sec:ape_dataset}.

\paragraph{RLHF-based Fine-tuning.}
We construct the preference datasets from the training splits to first train the reward model, followed by policy optimization using PPO guided by the reward model. Both reward model and PPO training are based on \texttt{Qwen2.5-14B-Instruct}, a strong decoder-only LLM that has demonstrated effectiveness across various NLP tasks including relation extraction. Our implementation builds upon the \texttt{OpenRLHF} package. Detailed information about the dataset construction, hyperparameters, and training costs are provided in Appendix \ref{sec:ape_training_detail}.

\paragraph{Evaluation Metrics.}
For all benchmarks, we evaluate model performance on the test sets using precision (P), recall (R), and micro-F1, which are widely adopted for relation extraction. In particular, to assess the model’s ability to discern relation existence, we pay special attention to the hallucination rate (HR) in addition to overall metrics.

\paragraph{Baselines.}
We compare DEPTH against the following state-of-the-art LLM-based relation extraction baselines:
\begin{itemize}[leftmargin=12pt,topsep=0pt,itemsep=0pt]
\item \textbf{QA4RE} \cite{zhang2023aligning} and \textbf{SUMASK} \cite{li2023revisiting}: zero-shot methods that reformulate relation extraction as question answering via prompt design;
\item \textbf{GPTRE} \cite{wan2023gpt}: an ICL method that enhances relation extraction by incorporating task-aware demonstration retrieval and goal-labeled reasoning;
\item \textbf{InstructUIE} \cite{wang2023instructuie}: an end-to-end information extraction framework that constructs expert-written instructions to fine-tune LLMs.
\end{itemize}
More details about these baselines are provided in Appendix~\ref{sec:ape_baselines}. We also include four straightforward baselines built on \texttt{Qwen2.5-14B-Instruct}. \textbf{Qwen-ZS} is a zero-shot setup that uses the same vanilla prompt as Section~\ref{sec:set_up}. \textbf{Qwen-ICL} is an in-context setup that employs the \textit{Grounding} prompt but omits the dependency-parsing component. \textbf{Qwen-SFT} performs supervised fine-tuning on the in-context \textit{Grounding} prompt, and \textbf{Qwen-RLHF} follows a standard RLHF pipeline with conventional reward model training and PPO-based policy optimization. For fair comparison, both Qwen-SFT and Qwen-RLHF adopt the same in-context prompting format as Qwen-ICL.

\subsection{Main Results}
\paragraph{DEPTH effectively reduces hallucinations in relation existence detection (Q1).}  
As shown in Table~\ref{tab:main-exist}, DEPTH achieves significantly lower hallucination rates across all benchmarks, with an average HR of only 7.9\%. In contrast, even the second-best model, InstructUIE, still suffers from an average HR of 18.4\%, underscoring DEPTH’s ability to handle \texttt{NO-RELATION} cases with significantly greater reliability.

\begin{table}[t]
  \setlength{\tabcolsep}{5pt}
  \centering
  \resizebox{\linewidth}{!}{
  \begin{tabular}{lllccccccccc}
    \toprule
    \multicolumn{2}{c}{Method} & Backbone & TACRED & TACREV & Re-TACRED & NYT11 & SciERC & FOBIE & DDI & SemEval & Avg $\downarrow$\\
    \midrule
    \multirow[t]{3}{*}{\emph{Zero-shot}}
      & QA4RE   & \texttt{Text-davinci-003} & 40.6 & 41.2 & 32.7 & 62.9 & 48.3 & 41.5 & 66.4 & 55.9 & 48.7 \\
      & Qwen-ZS & \texttt{Qwen2.5-14B-Instruct} & 55.6 & 55.7 & 57.9 & 100.0 & 97.0 & 99.8 & 83.8 & 95.4 & 80.7 \\
      & SUMASK  & \texttt{GPT-3.5-turbo-0301} & 20.4 & 24.9 & 26.2 & 65.3 & 44.2 & 31.8 & 76.2 & 40.8 & 41.2 \\
    \addlinespace[2pt]
    \multirow[t]{2}{*}{\emph{ICL}}
      & GPTRE    & \texttt{Text-davinci-003} & 27.9 & 31.6 & 38.4 & 46.8 & 31.0 & 36.5 & 47.8 & \underline{12.7} & 34.1 \\
      & Qwen-ICL & \texttt{Qwen2.5-14B-Instruct} & 28.7 & 27.7 & 33.6 & 92.9 & 68.9 & 52.9 & 55.6 & 68.3 & 53.6 \\
    \addlinespace[2pt]
    \multirow[t]{4}{*}{\emph{Fine-tuning}}
      & Qwen-SFT & \texttt{Qwen2.5-14B-Instruct} & 13.4 & 11.8 & 21.4 & 65.1 & 56.8 & 62.5 & 52.5 & 56.4 & 42.5 \\
      & InstructUIE & \texttt{Flan-T5-11B} & 20.2 & 20.2 & \underline{19.3} & \underline{16.9} & \underline{25.1} & \underline{23.1} & \underline{9.3} & 13.0 & \underline{18.4} \\
      & Qwen-RLHF & \texttt{Qwen2.5-14B-Instruct} & \underline{5.3} & \underline{3.1} & 20.3 & 45.3 & 31.4 & 28.6 & 13.4 & 37.2 & 23.1 \\
      & \textbf{DEPTH (ours)} & \texttt{Qwen2.5-14B-Instruct} & \textbf{3.6} & \textbf{2.6} & \textbf{9.7} & \textbf{13.5} & \textbf{12.6} & \textbf{2.7} & \textbf{7.6} & \textbf{11.1} & \textbf{7.9}\\
    \bottomrule
  \end{tabular}}
  \caption{\textbf{(Q1)} Hallucination rate (lower is better) for relation existence detection across eight datasets. \textbf{Bold} indicates the best performance and \underline{underlined} indicates the second-best.}
  \label{tab:main-exist}
\end{table}

\begin{table}[t]
  \setlength{\tabcolsep}{5pt}
  \centering
  \resizebox{\linewidth}{!}{
  \begin{tabular}{lllccccccccc}
    \toprule
    \multicolumn{2}{c}{Method} & Backbone & TACRED & TACREV & Re-TACRED & NYT11 & SciERC & FOBIE & DDI & SemEval & Avg $\uparrow$\\
    \midrule
    \multirow[t]{3}{*}{\emph{Zero-shot}}
      & QA4RE   & \texttt{Text-davinci-003} & 59.4 & 59.4 & 67.3 & 61.8$^\dagger$ & 51.7$^\dagger$ & 58.5$^\dagger$ & 52.6$^\dagger$ & 44.1 & 56.9 \\
      & Qwen-ZS & \texttt{Qwen2.5-14B-Instruct} & 45.9 & 46.9 & 45.1 & \underline{64.1} & 17.3 & 16.5 & 20.1 & 46.1 & 37.8 \\
      & SUMASK  & \texttt{GPT-3.5-turbo-0301} & 79.6 & 75.1 & 73.8 & 58.6$^\dagger$ & 55.8$^\dagger$ & 68.2$^\dagger$ & 31.5$^\dagger$ & 59.2$^\dagger$ & 62.7 \\
    \addlinespace[2pt]
    \multirow[t]{2}{*}{\emph{ICL}}
      & GPTRE    & \texttt{Text-davinci-003} & 72.1 & 68.4$^\dagger$ & 61.6$^\dagger$ & 56.5$^\dagger$ & \underline{69.0} & 63.5$^\dagger$ & 44.2$^\dagger$ & \underline{91.9} & 65.9 \\
      & Qwen-ICL & \texttt{Qwen2.5-14B-Instruct} & 69.4 & 72.0 & 68.5 & 31.2 & 43.7 & 58.2 & 49.2 & 59.2 & 56.4 \\
    \addlinespace[2pt]
    \multirow[t]{4}{*}{\emph{Fine-tuning}}
      & Qwen-SFT & \texttt{Qwen2.5-14B-Instruct} & 81.2 & 84.5 & \underline{78.1} & 47.0 & 51.9 & 50.7 & 51.5 & 61.5 & 63.3 \\
      & InstructUIE & \texttt{Flan-T5-11B} & 65.1$^\dagger$ & 66.3$^\dagger$ & 50.4$^\dagger$ & 61.0$^\dagger$ & 45.2 & 65.0$^\dagger$ & 83.0$^\dagger$ & 73.2 & 63.7 \\
      & Qwen-RLHF & \texttt{Qwen2.5-14B-Instruct} & \underline{84.2} & \textbf{90.4} & 77.9 & 54.0 & 67.7 & \underline{75.8} & \underline{83.1} & 79.6 & \underline{76.6} \\
      & \textbf{DEPTH (ours)} & \texttt{Qwen2.5-14B-Instruct} & \textbf{86.2} & \underline{90.0} & \textbf{87.7} & \textbf{72.9} & \textbf{76.3} & \textbf{94.3} & \textbf{84.2} & \textbf{95.3} & \textbf{85.9}\\
    \bottomrule
  \end{tabular}}
  \caption{\textbf{(Q2)} Micro-F1 (higher is better) on eight relation extraction datasets. Scores marked with $\dagger$ are reproduced due to unreported results in the original papers. Best and second-best scores are highlighted in \textbf{bold} and \underline{underlined}, respectively.}
  \label{tab:main-extract}
\end{table}

Compared with existing fine-tuning baselines, DEPTH demonstrates a clear advantage in hallucination mitigation. While Qwen-SFT achieves comparable performance on some news-domain datasets such as TACRED, its hallucination rate increases substantially on scientific and biomedical domains. More importantly, Qwen-RLHF, which follows a standard RLHF pipeline but does not incorporate dependency-aware simplification or causal reward modeling, still exhibits a markedly higher hallucination rate of 23.1\%. This highlights the necessity of integrating structural guidance and disentangled reward learning to guide LLMs toward more reliable predictions. A detailed ablation study on the contribution of each component in DEPTH is provided in Section~\ref{sec:ablation}.

Another notable observation is that zero-shot extraction approaches, such as Qwen-ZS, exhibit high HRs on scientific-domain benchmarks like SciERC and FOBIE. This may be attributed to their reliance on shallow lexical cues and limited contextual understanding, which leads to overprediction when faced with complex sentence structures or domain-specific terminology. While ICL methods such as Qwen-ICL achieve partial mitigation by encoding task-specific definitions, their performance varies considerably across datasets, indicating poor generalization. DEPTH addresses these limitations by integrating dependency-aware sentence simplification with causally guided fine-tuning, thereby ensuring robust and consistent extraction performance even in challenging scenarios.

\paragraph{DEPTH achieves an average improvement of 9.3\% in F1 score over existing LLM-based baselines (Q2).}
Table~\ref{tab:main-extract} summarizes the micro-F1 results across eight benchmarks, showing that DEPTH consistently outperforms all baselines with an average score of 85.9\%. This substantial gain demonstrates that effectively reducing hallucinations directly contributes to higher-quality relation extraction. Consistent with the findings in Table~\ref{tab:main-exist}, the largest improvements occur on the challenging scientific-domain benchmarks, where sentences are typically longer and more structurally complex, and thus more susceptible to hallucinated predictions by LLMs. This strong performance highlights DEPTH’s ability to generalize to complex, domain-specific text, making it well-suited for real-world applications such as enterprise document processing, where linguistic and conceptual complexity are common.

\begin{figure}[t]
  \centering
  \includegraphics[width=.6\linewidth]{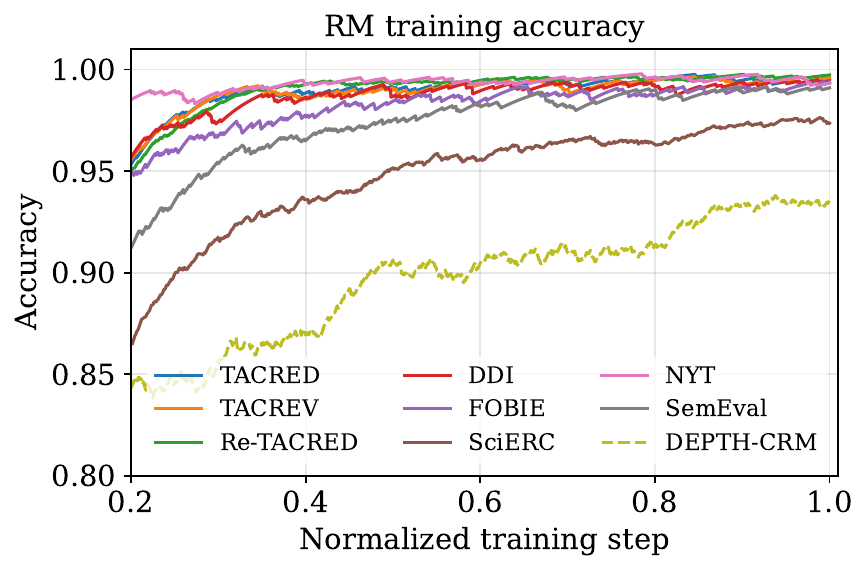}
  \caption{\textbf{(Q3)} Pairwise accuracy during RM training. DEPTH$-$CRM refers to the ablation variant where causal reward modeling is removed, details are provided in Section~\ref{sec:ablation}.}
  \label{fig:exp-rm}
\end{figure}

\begin{table}[t]
  \centering
  \begin{tabular}{llcccc}
    \toprule
    Domain & Benchmark & P $\uparrow$ & R $\uparrow$ & F1 $\uparrow$ & HR $\downarrow$\\
    \midrule
    \multirow{3}{*}{News} 
           & NYT11        & 79.1 & 81.3 & 80.2 & 18.7\\
           & Re-TACRED        & 81.9 & 96.9 & 88.8 & 3.1\\
           & \textbf{Average} & \textbf{80.5} & \textbf{89.1} & \textbf{84.5} & \textbf{10.9} \\
    \midrule
    \multirow{3}{*}{Scientific} 
           & SciERC        & 70.4 & 92.9 & 80.1 & 7.1 \\
           & FOBIE        & 94.4 & 96.2 & 95.3 & 3.8\\
           & \textbf{Average} & \textbf{82.4} & \textbf{94.5} & \textbf{87.7} & \textbf{5.5}\\
    \bottomrule
  \end{tabular}
  \caption{\textbf{(Q4)} Relation existence detection performance when a single DEPTH model is jointly trained on multiple datasets within the same domain and evaluated on each dataset's test set. This assesses whether DEPTH can serve as a general-purpose extractor without task-specific training.}
  \label{tab:main-cross}
\end{table}




\paragraph{DEPTH learns a high-quality reward model that is robust to spurious correlations (Q3).}  
Our causal factorization approach restructures each prompt–response pair in the preference dataset to extract the minimal sufficient information required for reward model training. This is implemented via template-based matching, ensuring that variations in guideline length or the inclusion of verbose explanations do not affect the factorization process. As a result, DEPTH effectively shields the reward model from common spurious factors such as prompt length, formatting artifacts, and sycophantic phrasing. Figure \ref{fig:exp-rm} reports the pairwise accuracy of the reward model throughout training across all benchmarks. Here, the x-axis denotes the normalized training steps, and the y-axis reflects the proportion of instances in which the chosen response is correctly assigned a higher reward than the rejected one. Our causal reward model achieves an average accuracy of 98.9\% across these tasks. Such consistency would be unattainable if the model were driven by spurious correlations.

\paragraph{DEPTH supports model sharing across datasets within similar domains (Q4).}  
In practical applications, training a separate task-specific LLM for each dataset is often prohibitively expensive, underscoring the need for a general-purpose relation extraction framework. To assess the cross-dataset applicability of DEPTH, we jointly train a single model on multiple datasets within the same domain and evaluate its performance on their individual test sets. Specifically, we select NYT11 and Re-TACRED to represent the news domain, and SciERC and FOBIE for the scientific domain. For relation types across datasets, we unify shared relations, while dataset-specific relations are preserved. The complete mapping rules are provided in Appendix~\ref{ape:unify}. Table \ref{tab:main-cross} reports the hallucination rates, demonstrating that the jointly trained DEPTH model remains competitive. These findings highlight DEPTH's ability to balance training efficiency with high extraction accuracy, making it highly applicable for real-world deployment where scalable and cost-effective solutions are essential.

\subsection{Ablation Studies}\label{sec:ablation}
To assess the contribution of each component, we further conduct ablation studies on the SciERC dataset. Detailed results are reported in Table \ref{tab:exp-ablation}.

\begin{table}[t]
  \centering
  \begin{tabular}{lcccc}
    \toprule
    Model     & P $\uparrow$ & R $\uparrow$ & F1 $\uparrow$ & HR $\downarrow$\\
    \midrule
    DEPTH               & \textbf{79.1} & \textbf{87.4} & \textbf{83.1} & \textbf{12.6}\\
    DEPTH$-$DP            & 71.0 & 78.1 & 74.4 & 21.9\\
    DEPTH$-$CRM           & 66.4 & 68.6 & 67.5 & 31.4\\
    DEPTH$-$Refinement    & 72.5 & 84.9 & 78.2 & 15.1\\
    \bottomrule
  \end{tabular}
  \caption{Ablation study results of relation existence detection on SciERC.}
  \label{tab:exp-ablation}
\end{table}

\begin{table}[t]
  \centering
  \begin{tabular}{lcccc}
    \toprule
    \multicolumn{1}{c}{Configuration} & P $\uparrow$ & R $\uparrow$ & F1 $\uparrow$ & HR $\downarrow$ \\
    \midrule

    \multicolumn{5}{l}{\textbf{(1) DEPTH with varying backbone scales}} \\
    \cmidrule(lr){1-5}
    \quad \texttt{Llama-3.2-3B-Instruct}    & 69.6 & 87.1 & 77.4 & 12.9 \\
    \quad \texttt{Llama-3.1-8B-Instruct}    & \underline{77.1} & \textbf{88.6} & \underline{82.5} & \textbf{11.4} \\
    \quad \texttt{Qwen2.5-14B-Instruct}     & \textbf{79.1} & \underline{87.4} & \textbf{83.1} & \underline{12.6} \\
    \midrule

    \multicolumn{5}{l}{\textbf{(2) Performance on \texttt{Qwen2.5-72B-Instruct}}} \\
    \cmidrule(lr){1-5}
    \quad Qwen-ZS    & \textbf{96.0} & 13.2 & 23.1 & 86.9 \\
    \quad Qwen-ICL   & 75.6 & 66.7 & 71.0 & 33.4 \\
    \quad Qwen-SFT   & \underline{93.2} & 89.2 & \underline{91.2} & 10.8 \\
    \quad Qwen-RLHF  & 90.8 & \underline{89.4} & 90.1 & \underline{10.6} \\
    \quad \textbf{DEPTH (ours)} & 91.8 & \textbf{94.5} & \textbf{93.1} & \textbf{5.5} \\
    \midrule

    \multicolumn{5}{l}{\textbf{(3) Performance of PLM-based methods}} \\
    \cmidrule(lr){1-5}
    \quad TriMF \cite{shen2021trigger}         & 89.8 & \textbf{93.8} & \textbf{91.8} & \textbf{6.2} \\
    \quad PL-Marker \cite{ye2022packed}     & \textbf{97.6} & 77.5 & 86.4 & 22.5 \\
    \bottomrule
  \end{tabular}
  \caption{Relation existence detection results on SciERC under three settings: (1) DEPTH with different LLM backbone sizes; (2) comparisons among LLM-based methods under the \texttt{Qwen2.5-72B-Instruct} backbone; (3) performance of representative PLM-based baselines.}
  \label{tab:exp-model-scope}
\end{table}

\paragraph{Impact of Dependency-Aware Simplification.}  
To examine the effect of sentence simplification and dependency-guided prompting on relation extraction, we remove all dependency parsing components from the DEPTH framework, resulting in a variant denoted as DEPTH$-$DP. Compared to the full model, DEPTH$-$DP yields a decrease of 8.1\% in precision, 9.3\% in recall, and 8.7\% in F1 score, thereby resulting in more hallucinations during extraction. This can be attributed to the syntactic complexity of many input sentences, which often causes LLMs to miss crucial relational cues during classification. These results suggest that leveraging SDPs for structural simplification effectively enhances the model's sentence understanding and leads to more accurate extraction.

\paragraph{Impact of Causal Reward Modeling.}
While previous results confirm that causal factorization enables the reward model to be robust against spurious correlations, we further investigate the impact of removing this component. Specifically, we ablate the \textit{Causal Reward Modeling} module, resulting in a variant denoted as DEPTH$-$CRM. As expected, this modification leads to a substantial performance decline. Notably, the hallucination rate rises to over 31\%. Figure \ref{fig:exp-rm} illustrates the training dynamics of the reward model by plotting its pairwise accuracy over training steps. Without applying causal factorization, the accuracy plateaus at 92.8\%, which is 4.3\% lower than that achieved with the factorized formulation. Such performance is inadequate to support effective RLHF fine-tuning. These findings highlight the critical importance of incorporating \textit{Causal Reward Modeling} into DEPTH, particularly in ICL scenarios where prompts may contain large amounts of irrelevant or confounding information.

\paragraph{Impact of Global Consistency Refinement.}  
The DEPTH$-$Refinement row in Table \ref{tab:exp-ablation} corresponds to the ablation variant where the \textit{Refinement} module is removed. Compared to the full DEPTH model, we observe consistent performace drops, indicating that the \textit{Refinement} module provides meaningful improvements across all metrics. These results demonstrate that the \textit{Refinement} step successfully verifies and self-corrects the predictions from the \textit{Grounding} module, thereby enhancing overall extraction quality.

\subsection{Sensitivity Analysis}
In this section, we assess DEPTH’s robustness to (1) dependency parser errors and (2) different LLM backbones.

\begin{promptbox}[float*=th,floatplacement=th]{Case Study: Recovery a Missing Relation via \textit{Refinement} in SciERC}
\textbf{Question.} Format the Prompt for the \textit{Refinement} Module in Appendix \ref{sec:ape_template} with Real Data:

\begin{snippetbox}
Sentence: The unique properties of tree-adjoining grammars (TAG) present a challenge for the application of \textbf{TAGs} beyond the limited confines of syntax, for instance, to the task of \textbf{semantic interpretation} or automatic translation of natural language .

\medskip
\begin{tabular}{@{}ll@{\hspace{1.5em}}ll@{}}
Entity 1: & \textbf{TAGs} & Entity 2: & \textbf{semantic interpretation}
\end{tabular}

\medskip
Dependency Parsing Information: The sentence is organized around the main predicate ``present'' as the root verb. The direct object of ``present'' is ``a challenge,'' and this challenge is elaborated by a for-PP (``for the application …'') that attaches to the verb and names what the challenge concerns: \textbf{``the application'' is complemented by ``of TAGs''} and further expanded by ``beyond the limited confines of syntax,'' where ``limited'' modifies ``confines'' and ``of syntax'' completes that noun. \textbf{Within this ``application'' domain, additional direction is given via ``to the task …,'' whose head ``task'' takes an of-PP (``of semantic interpretation'')} and a coordinated alternative (``or automatic translation''), with ``automatic'' modifying ``translation'' and ``of natural language'' dependent on ``translation.''

\medskip
All possible relations are: [\texttt{USED-FOR}, \texttt{FEATURE-OF}, \texttt{HYPONYM-OF}, \texttt{PART-OF}, \texttt{EVALUATE-FOR}, \texttt{COMPARE}, \texttt{CONJUNCTION}, \texttt{NO-RELATION}]

\medskip
\fcolorbox{red}{gray!10}{Candidate Relation for Entity 1 and Entity 2: \textbf{\texttt{Fail}}}

\medskip
All Candidate Relations for the sentence: [[\texttt{TAGs}, \texttt{semantic interpretation}, \texttt{Fail}], [\texttt{TAGs}, \texttt{syntax}, \texttt{NO-RELATION}], [\texttt{semantic interpretation}, \texttt{syntax}, \texttt{NO-RELATION}]]
\end{snippetbox}

\textbf{Answer:}
\begin{snippetbox}
\fcolorbox{red}{gray!10}{\{ "relationship": "USED-FOR" \}}

\medskip
The candidate relation for this pair is missing (\texttt{Fail}). From the dependency description, \textbf{``TAGs''} appears as the complement in \textbf{``the application of TAGs''}, while \textbf{``semantic interpretation''} heads the of-PP in \textbf{``the task of semantic interpretation''}. The sentence states that the application of TAGs extends to the task of semantic interpretation, which expresses a purpose/target of applying TAGs. This supports the asymmetric relation \texttt{USED-FOR}, i.e., \textbf{TAGs are used for semantic interpretation}.
\end{snippetbox}

\end{promptbox}

\paragraph{Dependency Parser Errors.}
DEPTH tolerates parser errors through its two-tiered modular design. First, dependency parsing outputs from \texttt{spaCy} serve as soft guidance, helping the LLM simplify sentences while preserving entity-pair information. Second, when the simplified context and its associated dependency description are insufficient for the \textit{Grounding} module to produce a valid relation label in the required output format for a given entity pair, it returns a \texttt{Fail} response. In such cases, the \textit{Refinement} module leverages the full sentence structure and global relational context to recover the missing predictions via its omission check.

To illustrate this correction mechanism, we present a case study from SciERC in which the \textit{Grounding} module fails to predict the relation between \textit{``TAGs''} and \textit{``semantic interpretation''}. By reasoning over the complete dependency structure of the original sentence, the \textit{Refinement} module correctly identifies that the relation between them is \texttt{USED-FOR}. Across all test instances, such failures occurred only four times, and each was successfully corrected by the \textit{Refinement} module.

\paragraph{LLM Sensitivity.}
To evaluate DEPTH’s robustness to backbone choices, we conduct experiments with \texttt{Llama-3.2-3B-Instruct}, \texttt{Llama-3.1-8B-Instruct}, and \texttt{Qwen2.5-} \texttt{14B-Instruct}, and further compare against both prompting-based and fine-tuning baselines under a much larger backbone, \texttt{Qwen2.5-} \texttt{72B-Instruct}. Table~\ref{tab:exp-model-scope} reports the corresponding relation existence detection results on SciERC.

A key observation is that scaling up the backbone alone does \emph{not} reliably resolve hallucinations. Even with \texttt{Qwen2.5-72B-Instruct}, simple prompting baselines still suffer from severe overprediction: Qwen-ZS exhibits an HR of 86.9\%, and Qwen-ICL, while mitigating hallucinations, still remains at 33.4\% HR. In contrast, fine-tuned baselines on the same 72B backbone reduce HR to around 10\%, suggesting that additional alignment beyond prompting is necessary for reliable \texttt{NO-RELATION} detection.

Notably, DEPTH is highly effective even with lightweight LLMs. For instance, applying DEPTH to a 3B backbone already achieves an HR of 12.9\%, which is comparable to strong fine-tuned baselines on a 72B backbone. Furthermore, DEPTH continues to improve performance even when applied to a 72B backbone, achieving the best HR (5.5\%) and F1 (93.1\%) in Table~\ref{tab:exp-model-scope}. Overall, these results indicate that DEPTH’s gains stem from its dependency-aware grounding and refinement design, rather than merely increasing model size.

\section{Related Work}
\subsection{Relation Extraction}
Relation extraction evolves from early supervised formulations that treat the task as classification over handcrafted features \cite{riloff1993automatically,kambhatla2004combining} and kernel methods \cite{lodhi2002text,zelenko2003kernel}. A prominent line of work leverages syntactic structure, in particular shortest dependency paths, to capture the minimal relational context between entities \cite{culotta2004dependency,bunescu2005shortest,fundel2007relex}. Semi-supervised and bootstrapping approaches have also been widely explored to reduce annotation cost \cite{brin1998extracting,agichtein2000snowball,etzioni2008open}. More recently, advances in deep neural networks and pre-trained language models (PLMs) have substantially improved the RE performance \cite{miwa2016end,nayak2020effective,yuan2021relation}.

LLMs have been extensively applied to relation extraction via data augmentation \cite{xu2023unleash}, zero-shot prompting \cite{zhang2023aligning,li2023revisiting}, in-context learning \cite{wan2023gpt}, and supervised fine-tuning \cite{cabot2021rebel}. Recent definition-driven approaches operate directly from natural-language relation specifications, and train lightweight NLI-style classifiers \cite{zhou2024grasping}. \citet{liu2024unleashing} leverages self-prompting to automatically generate diverse demonstrations for retrieval-augmented ICL. In addition to decoder-based LLMs, encoder-centric pipelines such as ITER \cite{hennen2024iter} jointly perform named entity recognition (NER) and RE with iterative transformer decisions, while unified generation frameworks like UIE \cite{lu2022unified} target general-purpose information extraction. InstructUIE \cite{wang2023instructuie} further improves end-to-end extraction through instruction tuning. Beyond sentence-level RE, document-level relation extraction \cite{jia2019document,xue2024autore} aims to infer relations that may require multi-sentence evidence aggregation, and has been studied with dedicated benchmarks such as DocRED \cite{yao2019docred} and DWIE \cite{zaporojets2021dwie}.

However, a key practical gap is reliably returning \texttt{NO-RELATION}, which is essential to avoid spurious edges in production-scale knowledge graphs. \citet{rogulsky2024effects} shows that hallucinations in synthetic training data can substantially harm recall, and that consistency-based filtering can improve quality. Our work advances LLM-based RE by explicitly modeling relation non-existence and evaluating hallucination via HR, coupling structured grounding with reward-guided training to discourage unwarranted relation assertions.

\subsection{Reward Hacking}
Reward hacking \cite{amodei2016concrete,gao2023scaling,skalse2022defining,eisenstein2023helping,wang2025beyond} arises when a reward model assigns high scores to outputs based on irrelevant or weakly correlated features, rather than truly aligning with the intended objectives. In the context of LLM alignment, typical sources of reward hacking include length \cite{dubois2024length}, formatting \cite{chen2024odin}, sycophancy \cite{perez2023discovering}, and superficial conceptual matches \cite{zhou2023explore}. In recent years, several studies have aimed to mitigate reward hacking, with one prominent direction focusing on causal analyses of its underlying mechanisms. For example, RMM \cite{liu2025rrm} introduces a data augmentation strategy grounded in a causal framework of reward model training to improve learning quality. \citet{ovinnikov2024learning} and \citet{wang2025beyond} incorporate causal invariance principles by introducing regularization constraints during training to encourage reward functions that are robust to spurious correlations. CAA \cite{xia2024aligning} treats the reward model as an instrumental variable to causally intervene on the LLM, effectively removing biases induced by confounding factors. Instead, our method employs structured causal factorization to directly disentangle reward-relevant and reward-irrelevant components in the input signals, which allows for the construction of a robust reward model without the need for extra training efforts.

\section{Limitations and Discussion}\label{sec:limitations}
\paragraph{RLHF training cost and practical trade-offs.}
While DEPTH demonstrates strong performance in hallucination-resistant relation extraction, its reliance on RLHF introduces higher computational overhead compared to zero-shot or ICL methods. As reported in Table~\ref{tab-ape:training_time}, reward model training averages 3.6 hours per dataset, and PPO fine-tuning requires approximately 4.5 days in our experiments. Several practical trade-offs are available under resource constraints. One approach is to reduce the backbone size. Encouragingly, DEPTH maintains robust performance even with lightweight LLMs (see Table~\ref{tab:exp-model-scope}). Furthermore, since we do not modify the model architecture and rely only on standard policy optimization, DEPTH is fully compatible with parameter-efficient fine-tuning (PEFT) techniques such as LoRA~\cite{hu2022lora}, which can significantly reduce memory and compute requirements during RLHF.

\paragraph{Extension to document-level relation extraction.}
The current formulation of DEPTH focuses on sentence-level relation extraction under the assumption that all entity mentions are provided in advance. Extending it to the more challenging document-level setting presents two primary challenges:
\begin{enumerate*}[label=(\arabic*)]
    \item accurately identifying all entity mentions within a document, and 
    \item resolving coreferences to link mentions that refer to the same real-world entity across sentences.
\end{enumerate*}
A straightforward approach is to prepend a ZS/ICL-based LLM pipeline for NER and coreference resolution, and then apply DEPTH to the resulting candidate entity pairs. However, errors from these upstream components may propagate and degrade end-to-end reliability. Alternatively, one could fine-tune dedicated NER and coreference models within the pipeline, but this would increase system complexity and training overhead. A more promising direction is end-to-end joint modeling, where a single LLM is prompted or fine-tuned to perform NER, coreference resolution, and relation extraction simultaneously.

\paragraph{Beyond Closed-Set Relation Extraction.}
A natural direction for future research is to extend DEPTH to open-set settings, where novel relation types may emerge. DEPTH’s strong capability in reliably identifying \texttt{NO-RELATION} instances provides a practical entry point for such an extension: one can first apply DEPTH in its standard closed-set mode and subsequently invoke an open-relation extraction module exclusively on the pairs predicted as \texttt{NO-RELATION}. For instance, the \textit{Refinement} prompt could be extended to jointly output the best closed-set label and a concise natural-language description of the relational evidence. The descriptions generated from \texttt{NO-RELATION} instances could then be clustered to propose candidate novel relation types, which may be further validated and incorporated into the schema. We leave a systematic evaluation of open-set relation discovery and ontology expansion to future work.

\paragraph{Relation to supervised RE methods.}
DEPTH is primarily motivated by the challenge of hallucination in generative, LLM-based relation extraction. Meanwhile, a major line of RE research relies on discriminative pre-trained language models (PLMs) fine-tuned for relation extraction, which remain strong baselines for sentence-level tasks. Although PLM-based methods are not the primary focus of this work, we include representative PLM baselines in Table~\ref{tab:exp-model-scope} for completeness. The results show that DEPTH achieves the best overall performance, while PLM baselines are also competitive and substantially outperform non-fine-tuned LLM baselines such as Qwen-ZS and Qwen-ICL. This is consistent with prior observations that well-tuned PLMs often provide more reliable extraction than zero-shot or ICL-based LLM prompting~\cite{li2023evaluating,zhao2024comprehensive,xu2024large}. 

\section{Conclusion and Future Work}
In this paper, we present DEPTH, a methodology designed to mitigate the hallucination phenomenon in LLM-based relation extraction, where models tend to erroneously assign relations to entity pairs labeled as \texttt{NO-RELATION}. In particular, DEPTH adopts a two-tiered refinement framework: the \textit{Grounding} module enhances local prediction accuracy via dependency-aware sentence simplification and RLHF fine-tuning guided by a causally trained, robust reward model; the \textit{Refinement} module performs sentence-level integration and applies global semantic constraints to verify and self-correct the outputs from the \textit{Grounding} stage. Extensive experiments on a suite of well-established benchmarks validate the effectiveness of DEPTH in improving both extraction precision and robustness to hallucination. Future research directions include integrating NER into the pipeline to construct a fully end-to-end relation extraction framework, extending DEPTH to more challenging settings such as document-level extraction and open-set RE, and applying it to retrieval-augmented generation (RAG) systems by constructing reliable knowledge graphs from unstructured text to enhance evidence retrieval.

\bibliographystyle{unsrtnat}
\bibliography{reference}


\appendix

\newpage

\section{Dataset Descriptions}\label{sec:ape_dataset}
We conduct experiments on eight datasets for evaluating the relation extraction capabilities of LLMs. To assess the generalizability of DEPTH across different domains, we select datasets spanning the news domain (TACRED, TACREV, Re-TACRED, NYT11), the scientific domain (SciERC, FOBIE), a biomedical-domain dataset (DDI), and a general-domain benchmark (SemEval).

\begin{itemize}[leftmargin=12pt,topsep=0pt,itemsep=0pt]
\item \textbf{TACRED} is a widely used benchmark for relation extraction, known for its rich set of relation types and high-quality annotations. It contains over 100,000 labeled instances derived from newswire and web text, covering 42 distinct relation types.
\item \textbf{TACREV} improves upon TACRED by correcting annotation errors, particularly in the dev and test sets. The dataset also includes grouped analyses of these errors to facilitate more reliable evaluation.
\item \textbf{Re-TACRED} is a reconstructed version of TACRED, designed to address biases and imbalances in the original dataset. Compared to TACRED, it introduces expanded relation types and entity categories, reflecting more diverse linguistic contexts and application scenarios.
\item \textbf{NYT11} is a supervised dataset constructed by \citet{hoffmann2011knowledge} through aligning Freebase relations to New York Times articles. It defines 24 valid relation types plus a \texttt{NO-RELATION} class.
\item \textbf{SciERC} is a benchmark for scientific information extraction, containing entities and relations annotated from computer science paper abstracts. It covers key categories such as tasks, methods, materials, and metrics.
\item \textbf{FOBIE} supports relation extraction in the context of computer-assisted biomimetics. It consists of 1,500 manually annotated sentences from scientific biology literature, focusing on core conceptual relationships such as trade-offs and correlations.
\item \textbf{DDI} is a biomedical relation extraction dataset for drug--drug interaction extraction. It contains 792 texts selected from the DrugBank database and 233 Medline abstracts. Drug-drug interactions, including both pharmacokinetic and pharmacodynamic types, were annotated by expert pharmacists.
\item The \textbf{SemEval}-2010 Task 8 dataset includes 9 semantic relation types and a \texttt{NO-RELATION} class. It contains 10,717 sentences from a variety of domains, each annotated with entities and their relations. The dataset is known for its annotation quality and consistency, achieved through expert labeling and multi-round validation.
\end{itemize}

The statistics of these datasets are summarized in Table \ref{tab-ape:dataset_statistics}. For each dataset, we merge the train and dev sets to construct the preference dataset $\mathcal{D}$ used for training both the reward model and PPO policy. Specifically, consider a sentence $s$ and an entity pair $(e_1, e_2)$ with ground-truth relation $\rho$. Let $x$ denote the prompt template used to query the LLM, and let $\mathcal{R}$ be the predefined set of relation types. For every negative relation $\rho' \in \mathcal{R} \setminus \{\rho\}$, we construct a JSON-formatted preference example in $\mathcal{D}$ as follows:
\begin{promptbox}{Example JSON Entry of the Preference Dataset}
\begin{flushleft}
\{
\\ \quad ``question'': ``$x(s, e_1, e_2)$'',\\
\quad ``correct'': \{\\
\qquad ``relationship'': ``$\rho$'',\\
\qquad ``explanation'': ``LLM-generated rationale for predicting $\rho$''\\
\quad \},\\
\quad ``incorrect'': \{\\
\qquad ``relationship'': ``$\rho'$'',\\
\qquad ``explanation'': ``LLM-generated rationale for why $\rho'$ could be predicted''\\
\quad \}\\
\}
\end{flushleft}
\end{promptbox}

Specifically, the \textit{``question''} field is created by filling the sentence $s$, entity pair $(e_1, e_2)$, and other relevant information into the prompt template $x$. The \textit{``correct''} response contains the true relation $\rho$ and a rationale generated by the LLM, while the \textit{``incorrect''} response includes a negative relation $\rho'$ and a plausible explanation, also generated by the LLM, for why it might be chosen. At test time, we format all examples in the test set using the same prompt template $x$ to ensure consistency in evaluation.

\begin{table}[ht]
  \centering
  \begin{tabular}{lccccc}
    \toprule
    Benchmark & Train & Dev & Test & Relations & \texttt{NO-RELATION} \\
    \midrule
    TACRED & 68,124 & 22,631 & 15,509 & 42 & 79.4\% \\
    TACREV & 68,124 & 22,631 & 15,509 & 42 & 79.4\% \\
    Re-TACRED & 58,196 & 19,584 & 13,148 & 40 & 63.2\% \\
    NYT11 & 335,843 & 37,010 & 1,450 & 25 & 64.1\% \\
    SciERC & 16,872 & 2,033 & 4,088 & 8 & 90.2\% \\
    FOBIE & 6,913 & 785 & 801 & 3  & 73.2\% \\
    DDI & 18,779 & 7,244 & 5,754 & 5 & 83.0\% \\
    SemEval & 6,507 & 1,496 & 2,717 & 10 & 17.4\% \\
    \bottomrule
  \end{tabular}
  \caption{Statistics of the datasets used in our experiments.}
  \label{tab-ape:dataset_statistics}
\end{table}

\section{Baseline Descriptions}\label{sec:ape_baselines}
In this section, we provide additional details on the baseline methods compared in our experiments.
\begin{itemize}[leftmargin=12pt,topsep=0pt,itemsep=0pt]
    \item \textbf{QA4RE} reformulates relation extraction as multiple-choice question answering to better match the predominant QA-style tasks seen during instruction tuning. Given a sentence and an entity pair, it asks the LLM a QA question whose options correspond to relation labels, and selects the option returned by the model as the predicted relation.
    \item \textbf{SUMASK} is a zero-shot prompting framework that adopts a summarize-and-ask strategy. It first prompts the LLM to summarize the input into a relation-focused form and then queries the model again in a QA-like format for the relation label, aiming to reduce irrelevant context and improve robustness compared to a single vanilla prompt.
    \item \textbf{GPTRE} is an ICL approach for relation extraction. It improves ICL by retrieving demonstrations using task-aware representations that emphasize entity and relation information, and augmenting demonstrations with gold-label-induced reasoning to make the input--label mapping more explicit for the LLM.
    \item \textbf{InstructUIE} is an instruction-tuned framework for unified information extraction. It converts RE tasks into a text-to-text format with expert-written instructions and performs supervised instruction tuning on a multi-task benchmark, enabling a single model to generalize across tasks under both supervised and zero-shot settings.
\end{itemize}

\section{Training Details}\label{sec:ape_training_detail}
\subsection{Reward Model Training}\label{sec:ape_rm_training}
\paragraph{Loss Function}  
Let the reward model be denoted as $r_{\phi}(x, y)$. Given a preference dataset $\mathcal{D}$, we train $r_{\phi}$ using the following loss function, which is commonly adopted in the RLHF literature \cite{ahmadian2024back}:
\begin{equation}
    \mathcal{L}_{RM} = - \frac{1}{| \mathcal{D}|} \mathbb{E}_{(x, y^+, y^-) \in \mathcal{D}} \log \left( \sigma \left( r_{\phi}(x,y^+) - r_{\phi}(x,y^-)\right) \right).
\label{eq-ape:rm_loss}
\end{equation}
Here, $(x, y^+, y^-)$ is a triplet from $\mathcal{D}$, where $y^+$ is the preferred response and $y^-$ is the dispreferred one. $\sigma(\cdot)$ denotes the sigmoid function.

\paragraph{Implementation of Causal Factorization}
We implement causal factorization by splitting the \textit{Grounding} prompt $x$ and the corresponding LLM output $y$ into reward-relevant and reward-irrelevant parts, i.e., $x=(x_1,x_2)$ and $y=(y_1,y_2)$. Following the definition in the main text, $x_1$ is the minimal prompt component sufficient for reward prediction, consisting of the task definition, the simplified sentence, the target entity pair, and a natural-language description of their shortest dependency path. Concretely, in our implementation, $x_1$ is instantiated using the following fixed template:
\begin{promptbox}{Reward-relevant Prompt}
You are an information extraction system. Your goal is to determine the relationship between two given entities based on the provided sentence.

\begin{snippetbox}
Sentence: $s$ \\
Entity 1: $e_1$
\quad Entity 2: $e_2$ \\
Dependency Parsing Information: DP\_INFO \\
All possible relations are: $\mathcal{R}$
\end{snippetbox}

Answer:
\end{promptbox}

The full prompt used by the \textit{Grounding} module is provided in Appendix~\ref{sec:ape_template}, where the exact text span corresponding to $x_1$ is highlighted in blue, and the remaining components, such as in-context examples and auxiliary extraction guidelines, are highlighted in gray and assigned to $x_2$. In other words, we obtain $x_2$ by taking the full \textit{Grounding} prompt and removing the $x_1$ span defined above.

Similarly, the reward-relevant response $y_1$ is strictly confined to the predicted relation label $\hat{\rho}$ formatted in a fixed JSON field, as shown below:
\begin{promptbox}{Reward-relevant Response}
Answer:
\begin{snippetbox}
\{ "relationship": "$\hat{\rho}$" \}
\end{snippetbox}
\end{promptbox}

Any additional text generated by the LLM, such as explanatory rationales or formatting artifacts, is assigned to $y_2$. This factorization is applied at the input–output level and is fixed \emph{a priori} through predefined templates, ensuring a clean separation between reward-relevant and reward-irrelevant signals during reward modeling. Accordingly, we replace Eq. (\ref{eq-ape:rm_loss}) with its causally factorized variant:
\begin{equation}
    \mathcal{L}_{RM} = - \frac{1}{| \mathcal{D}|} \mathbb{E}_{(x_1, y_1^+, y_1^-) \in \mathcal{D}} \log \left( \sigma \left( r_{\phi}(x_1,y_1^+) - r_{\phi}(x_1,y_1^-)\right) \right).
\label{eq-ape:crm_loss}
\end{equation}

\paragraph{Hyperparameters}  
Our implementation is based on the \texttt{OpenRLHF} (v0.8.5) library. Aside from the causal factorization components, we make no modifications to the original codebase. The hyperparameters specifically used in our experiments are summarized in Table \ref{tab-ape:training-hyper}.

\begin{table}[t]
  \centering
  \small
  \begin{minipage}[t]{0.56\linewidth}
    \centering
    \begin{tabular}{lcc}
      \toprule
      Hyperparameters & RM & PPO \\
      \midrule
      bf16 & True & True \\
      zero stage & 3 & 3 \\
      batch size & 64 & 64 \\
      num epoch/episodes & 1 & 1 \\
      learning rate & $9\times 10^{-6}$ & -- \\
      max input length & 1024 & -- \\
      actor lr & -- & $5\times 10^{-7}$ \\
      critic lr & -- & $9\times 10^{-6}$ \\
      prompt max length & -- & 4096 \\
      generation max length & -- & 1024 \\
      rollout batch size & -- & 512 \\
      training epochs & -- & 1 \\
      init kl coeff & -- & 0.01 \\
      normalize reward & -- & True \\
      GAE $\lambda$ & -- & 1.0 \\
      GAE $\gamma$ & -- & 1.0 \\
      temperature & -- & 1.0 \\
      optimizer & -- & Adam \\
      \bottomrule
    \end{tabular}
    \caption{Hyperparameters for RM and PPO training.}
    \label{tab-ape:training-hyper}
  \end{minipage}\hfill
  \begin{minipage}[t]{0.40\linewidth}
    \centering
    \begin{tabular}{lcc}
      \toprule
      Benchmark & RM (h) & PPO (d) \\
      \midrule
      TACRED    & 4.7 & 6.0 \\
      TACREV    & 4.6 & 5.9 \\
      Re-TACRED & 4.6 & 5.7 \\
      NYT11     & 5.1 & 6.0 \\
      SciERC    & 2.6 & 3.1 \\
      FOBIE     & 0.6 & 1.7 \\
      DDI       & 3.7 & 2.6 \\
      SemEval   & 2.7 & 4.6 \\
      \midrule
      Average   & 3.6 & 4.5 \\
      \bottomrule
    \end{tabular}
    \caption{Training time per dataset.}
    \label{tab-ape:training_time}
  \end{minipage}
\end{table}

\paragraph{Training Costs}
Reward model training is conducted on a single compute node equipped with 8 NVIDIA H20 GPUs (each with 96GB VRAM), 192 CPU cores, and 1.8 TB RAM. The actual training time varies by dataset due to differences in data scale. Table \ref{tab-ape:training_time} summarizes the approximate training times for each dataset.

\subsection{PPO Training}
\paragraph{Hyperparameters}  
We also use the \texttt{OpenRLHF} library to implement PPO. To accommodate our causal reward model, we apply causal factorization to the reward computation. Apart from this, we make no further modifications to the codebase. The hyperparameters are summarized in Table \ref{tab-ape:training-hyper}.

\paragraph{Training Costs}
PPO optimization is conducted using the same compute resources as reward model training. We employ \texttt{Ray}-based distributed training \cite{moritz2018ray}. Table \ref{tab-ape:training_time} reports the approximate training times for each dataset in our experiments.

\subsection{Cross-Dataset Relation Unification}\label{ape:unify}
To evaluate whether DEPTH can serve as a general-purpose extractor, we jointly train a single DEPTH model on multiple datasets within the same domain. Since different datasets may use different label names for semantically equivalent relations, we perform a lightweight label unification step before constructing the joint training data. Specifically, we apply a predefined mapping function to each training instance that
\begin{enumerate*}[label=(\arabic*), itemjoin={{; }}, itemjoin*={{; and }}]
    \item rewrites a small set of relation labels that are clearly equivalent across datasets into a unified target label space
    \item leaves all remaining dataset-specific relation labels unchanged
\end{enumerate*}.
This ensures that shared relations are learned under a consistent label name, while preserving dataset-specific relations.

For the news domain, we align NYT11 with Re-TACRED by mapping 13 NYT11 relation types to their corresponding counterparts in Re-TACRED. Relations not covered by this mapping are kept under their original NYT11 labels. For the scientific domain, we align FOBIE with SciERC by mapping \texttt{Not\_a\_TradeOff} to \texttt{NO-RELATION}, which serves as the default negative class in our framework, while leaving all other FOBIE relations unchanged. Table~\ref{tab:relation-mapping} summarizes the full set of mappings used in training data preprocessing.

\begin{table}[htbp]
    \centering
    \resizebox{.92\linewidth}{!}{
    \begin{tabular}{lll}
        \toprule
        Domain & Original Label (Source Dataset) & Unified Label (Target Dataset) \\
        \midrule
        \multirow{13}{*}{News} 
            & \texttt{/no\_relation} & \texttt{no\_relation} \\
            & \texttt{/business/company/advisors} & \texttt{org:top\_members/employees} \\
            & \texttt{/business/company/founders} & \texttt{org:founded\_by} \\
            & \texttt{/business/company/major\_shareholders} & \texttt{org:shareholders} \\
            & \texttt{/business/person/company} & \texttt{per:employee\_of} \\
            & \texttt{/people/person/children} & \texttt{per:children} \\
            & \texttt{/people/person/ethnicity} & \texttt{per:origin} \\
            & \texttt{/people/person/nationality} & \texttt{per:origin} \\
            & \texttt{/people/person/place\_lived} & \texttt{per:cities\_of\_residence} \\
            & \texttt{/people/person/place\_of\_birth} & \texttt{per:city\_of\_birth} \\
            & \texttt{/people/person/profession} & \texttt{per:title} \\
            & \texttt{/people/person/religion} & \texttt{per:religion} \\
            & \texttt{/people/deceased\_person/place\_of\_death} & \texttt{per:city\_of\_death} \\
        \midrule
        Scientific & \texttt{Not\_a\_TradeOff} & \texttt{NO-RELATION} \\
        \bottomrule
    \end{tabular}}
    \caption{Relation type unification used for cross-dataset training.}
    \label{tab:relation-mapping}
\end{table}

\newpage
\section{Prompt Templates}\label{sec:ape_template}
Below we present the complete set of prompt templates used in this work, using SciERC as an example. For other datasets, we retain the core structure of the prompts while replacing dataset-specific components such as the set of relation types. In addition, the prompts used for Qwen-ICL, Qwen-SFT, and Qwen-RLHF follow the same template as the \textit{Grounding} prompt, except that we remove the dependency-parsing-related input. For brevity, we do not separately list these prompts.

\paragraph{Vanilla Prompt.}
The first prompt template defines the problem setup and serves as the vanilla prompt used in Qwen-ZS. It consists of a single-sentence task definition and the required input data for relation extraction.

\begin{promptbox}{Vanilla Prompt for Problem Setup and Qwen-ZS}
Determine which relationship can be inferred from the given sentence and two entities. If no relationship exists, respond with \texttt{NO-RELATION}.

\begin{snippetbox}
Sentence: $s$

\medskip
\begin{tabular}{@{}l l@{}}
Entity 1: & $e_1$ \\
Entity 2: & $e_2$
\end{tabular}

\medskip
All possible relations: $\mathcal{R}$
\end{snippetbox}

\textbf{Answer:}
\end{promptbox}

\paragraph{Prompt for Converting Dependency Structures into Natural Language.}
This prompt guides the LLM to convert raw dependency parsing outputs from \texttt{spaCy} into natural language descriptions that preserve syntactic structure. The \textit{Grounding} module focuses on transforming the SDP between two entities, while the \textit{Refinement} module expands this process to the entire sentence by converting the corresponding dependency tree (DP tree).

\begin{promptbox}{Prompt for Converting Dependency Structures into Natural Language}
Convert the dependency parsing results into a natural language description that reflects the syntactic structure of the sentence.

\medskip
\textbf{— For \textit{Grounding} Module —}
Focus only on the dependency links between the entities, and ignore other parts of the sentence.

\begin{snippetbox}
Sentence: $s$

\medskip
\begin{tabular}{@{}l l@{}}
Entity 1: & $e_1$ \\
Entity 2: & $e_2$
\end{tabular}

\medskip
Shortest Dependency Path: \texttt{SDP}
\end{snippetbox}

\medskip
\textbf{— For \textit{Refinement} Module —}
The description should capture how key tokens are connected, including clause structures, modifiers, and dependencies between predicates and arguments.

\begin{snippetbox}
Sentence: $s$

\medskip
Dependency Parsing Results: \texttt{DP\_Tree}
\end{snippetbox}

\textbf{Answer:}
\end{promptbox}

\newpage
\paragraph{Prompt for Dependency-Aware Simplification.}
This prompt is divided into three parts: task definition, instructions, and real data. It guides the LLM to simplify sentences by leveraging the information from the SDP while preserving essential semantics.

\begin{promptbox}{Prompt for Dependency-Aware Simplification}
You are given:
\par - An original sentence that may contain extra information.
\par - Two target entities within that sentence.
\par - The shortest dependency path (SDP) connecting the two entities.

\medskip
\textbf{— Task Definition —}
\par Your task is to generate a concise, simplified version of the sentence that:
\par - Preserves the relationship between the two entities.
\par - Reflects the meaning conveyed by the SDP.
\par - Removes unnecessary details unrelated to the core relational context.
\par - Ensures grammatical correctness and clarity.

\medskip
\textbf{— Instructions —}
\par - Focus on the SDP and retain all critical words necessary to express the relation.
\par - Remove content that does not directly support the relation between Entity 1 and Entity 2.
\par - If simplification is not feasible or the SDP is uninformative, return the original sentence unchanged.
\par - Output your result in the following JSON format:

\begin{snippetbox}
\ttfamily
{"Simplified sentence": "..."}
\end{snippetbox}

\medskip
\textbf{— Real Data —}
\begin{snippetbox}
Original Sentence: $s$

\medskip
\begin{tabular}{@{}l l@{}}
Entity 1: & $e_1$ \\
Entity 2: & $e_2$
\end{tabular}

\medskip
Shortest Dependency Path: \texttt{SDP}
\end{snippetbox}

\textbf{Answer:}
\end{promptbox}

\paragraph{Prompts for the \textit{Grounding} and \textit{Refinement} Modules.}
Finally, we provide the prompt templates for the \textit{Grounding} and \textit{Refinement} modules, both of which are implemented as ICL prompts. Each prompt consists of the following components: task description, relation definitions, guidelines, examples, and real data. The relation definitions are identical for both templates. For datasets with official descriptions, we use their provided definitions. Otherwise, we generate initial definitions with LLMs and refine them manually. The guidelines highlight key instructions that we empirically summarized during the extraction process and enforce a standardized output format for the LLMs. Upon implementation, we randomly select samples from the training set for ICL. For simplicity, additional examples are omitted in the following illustrations. The number of in-context examples is empirically set to 10 for the \textit{Grounding} module and 5 for the \textit{Refinement} module. These in-context samples are excluded from the training process. 

Notably, for the \textit{Grounding} prompt used to train the causal reward model, we retain only the task definition and the real-data fields, as detailed in Section \ref{sec:ape_rm_training}.

\begin{promptbox}{Prompt for the \textit{Grounding} Module}
\small

\begin{blueblock}
You are an information extraction system. Your goal is to determine the relationship between two given entities based on the provided sentence.
\end{blueblock}

\medskip
\begin{grayblock}
\textbf{— Relation Definitions —}
\par SciERC contains 5 asymmetric semantic relation types, 2 symmetric relation types, plus a catch-all ``No-Relation''. 

Asymmetric relation types:
\par - ``USED-FOR'': B is used for A, B models A, A is trained on B, B exploits A, A is based on B.
\par - ``FEATURE-OF'': B belongs to A, B is a feature of A, B is under A domain.
\par - ``HYPONYM-OF'': B is a hyponym of A, B is a type of A.
\par - ``PART-OF'': B is a part of A, A includes B, incorporate B to A.
\par - ``EVALUATE-FOR'': A is an evaluation metric for B.
\par Symmetric relation types:
\par - ``COMPARE'': Opposite of conjunction, compare two models/methods, or list two opposing entities.
\par - ``CONJUNCTION'': A and B function in a similar role or use/incorporate with each other.
\par Generic relation type:
\par - ``NO-RELATION'': No relationship can be inferred between A and B.

\medskip
\textbf{— Important Guidelines —}
\par - Only rely on the provided relationship definitions without guessing or expanding them.
\par - If there is no clear or strong indication of a defined relation, respond with ``NO-RELATION''.
\par - Refer to the Dependency Parsing Information as semantic guidance for analyzing the context around the entities.
\par - For asymmetric relations, always assume $B \to A$.
\par - Provide the final answer in the following JSON format:

\begin{snippetbox}
{ "relationship": "..." }
\end{snippetbox}

\medskip
\textbf{— Examples —}

Example 1:
\begin{snippetbox}
Sentence: English is shown to be trans-context-free on the basis of coordinations of the respectively type.

\medskip
\begin{tabular}{@{}l l@{}}
Entity 1: & English \\
Entity 2: & coordinations
\end{tabular}

\medskip
Dependency Parsing Information: Entity 1 (‘English’) is the subject, depending on the verb ‘is shown’ with ‘is’. Entity 2 (‘coordinations’) is the object of the preposition ‘of’, depending on ‘of’ in the phrase ‘on the basis of coordinations’. There is no direct dependency between Entity 1 and Entity 2.

\medskip
All possible relations are: [\texttt{USED-FOR}, \texttt{FEATURE-OF}, \texttt{HYPONYM-OF}, \texttt{PART-OF}, \texttt{EVALUATE-FOR}, \texttt{COMPARE}, \texttt{CONJUNCTION}, \texttt{NO-RELATION}]
\end{snippetbox}

\textbf{Answer:}
\begin{snippetbox}
{ "relationship": "NO-RELATION" }
\end{snippetbox}

Example 2: ...

\medskip
\textbf{— Real Data —}
\end{grayblock}

\begin{blueblock}
Sentence: $s$

\medskip
\begin{tabular}{@{}l l@{}}
Entity 1: & $e_1$ \\
Entity 2: & $e_2$
\end{tabular}

\medskip
Dependency Parsing Information: DP\_INFO

\medskip
All possible relations are: $\mathcal{R}$
\end{blueblock}

\textbf{Answer:}
\end{promptbox}

\begin{promptbox}{Prompt for the \textit{Refinement} Module}
\small
You are an information extraction system. Your goal is to \textbf{confirm}, \textbf{correct}, or \textbf{add} the relationship between two given entities based on the provided sentence. \\

\textbf{— Relation Definitions —} \\
SciERC contains 5 asymmetric semantic relation types, 2 symmetric relation types, plus a catch-all ``No-Relation''...\\

\textbf{— Important Guidelines —} \\
- Review the Candidate Relation for Entity 1 and Entity 2.\\
- Decide whether to:\\
    \textbf{$\cdot \quad$ CONFIRM} the candidate relation. \\
    \textbf{$\cdot \quad$ CORRECT} the candidate relation.\\
    \textbf{$\cdot \quad$ ADD} if the candidate relation is missing (which, in this case, will be `Fail`, `None`, or `Null`).\\
- If you are not highly confident in correcting the candidate relation, prefer to CONFIRM rather than CORRECT it.\\
- If the candidate relation is missing, make sure to explicitly ADD the correct relation if one exists.\\
- Only rely on the provided relationship definitions without guessing or expanding them.\\
- If there is no clear or strong indication of a defined relation, respond with ``NO-RELATION''.\\
- Refer to the Dependency Parsing Information as semantic guidance for analyzing the context around the entities.\\
- Provide the final answer in the following JSON format:
\begin{snippetbox}
\{ "relationship": "..." \}
\end{snippetbox}

\textbf{— Examples —} \\
Example 1: CONFIRM\\
Sentence: The results of the experiment show that in most of the cases the cooccurrence statistics indeed reflect the semantic constraints and thus provide a basis for a useful disambiguation tool. \\
Entity 1: cooccurrence statistics\\
\quad Entity 2: semantic constraints\\
Dependency Parsing Information: The sentence is anchored by the main verb ``show,'' whose subject is ``the results,'' further specified by the prepositional phrase ``of the experiment.'' ... \\
All possible relations are: [``USED-FOR'', ``FEATURE-OF'', ``HYPONYM-OF'', ``PART-OF'', ``EVALUATE-FOR'', ``COMPARE'', ``CONJUNCTION'', ``NO-RELATION'']\\
Candidate Relation for Entity 1 and Entity 2: ``NO-RELATION''\\
All Candidate Relations for the sentence: [[``cooccurrence statistics'', ``semantic constraints'', ``NO-RELATION''], [``cooccurrence statistics'', ``disambiguation tool'', ``USED-FOR''], [``semantic constraints'', ``disambiguation tool'', ``NO-RELATION'']]\\

Answer:
\begin{snippetbox}
\{ "relationship": "NO-RELATION" \}
\end{snippetbox}

Example 2: CORRECT\\
...\\

\textbf{— Real Data —}
\begin{snippetbox}
Sentence: $s$ \\
Entity 1: $e_1$
\quad Entity 2: $e_2$ \\
Dependency Parsing Information: DP\_INFO \\
All possible relations are: $\mathcal{R}$\\
Candidate Relation for Entity 1 and Entity 2: cand\_rel\\
All Candidate Relations for the sentence: predicted\_rels 
\end{snippetbox}

\textbf{Answer:}
\end{promptbox}

\newpage
\section{Statistical Significance Tests}
\label{sec:significance}

We further conduct statistical significance tests for DEPTH’s improvements on SciERC.
Speficially, we consider two complementary perspectives: (1) whether DEPTH significantly reduces hallucinations on \texttt{NO-RELATION} instances $D_2$; and (2) whether it significantly improves end-to-end extraction performance measured by micro-F1 on the full test set $D$.

\subsection{McNemar’s Test for Hallucination Reduction on \texttt{NO-RELATION}}
\label{sec:mcnemar}

Recall that the hallucination rate (HR) is defined on the \texttt{NO-RELATION} subset $D_2$ as the fraction of instances where the model predicts any relation other than \texttt{NO-RELATION}.
To test whether DEPTH reduces hallucinations compared to a baseline, we apply McNemar’s test, which is suitable for paired binary outcomes.

\paragraph{Setup.}
We restrict evaluation to instances with gold label \texttt{NO-RELATION}. For each instance, we define a binary event:
\[
\textsc{Correct} \iff \hat{\rho}=\texttt{NO-RELATION},\qquad
\textsc{Wrong} \iff \hat{\rho}\neq\texttt{NO-RELATION},
\]
which corresponds to \emph{non-hallucination} vs.\ \emph{hallucination} on $D_2$.
We then construct a $2\times 2$ contingency table over paired outcomes of DEPTH and the baseline, and perform an exact two-sided McNemar’s test on the discordant pairs.

\paragraph{Results.}
Table~\ref{tab:mcnemar_scierc} reports the contingency table for DEPTH vs.\ InstructUIE on SciERC $D_2$.
McNemar’s test yields a two-sided $p$-value of $5.14\times 10^{-18}$, indicating that DEPTH’s hallucination reduction is statistically significant under the conventional 0.05 threshold (i.e., $p<0.05$).

\subsection{Paired Bootstrap Test on micro-F1}
\label{sec:bootstrap_f1}

Beyond \texttt{NO-RELATION} hallucination, we also test whether DEPTH’s overall improvement in micro-F1 on the full test set $D$ is statistically significant.
Since micro-F1 is computed at the corpus level, we use a paired bootstrap test by resampling test instances with replacement.

\paragraph{Setup.}
Given aligned predictions from DEPTH and a baseline on $D$, we repeatedly resample the instances with replacement (paired across models), compute micro-F1 for each model, and record the difference
$\Delta=\text{F1}_{\text{DEPTH}}-\text{F1}_{\text{baseline}}$.
We run 10{,}000 resamples and report the 95\% percentile confidence interval (CI) of $\Delta$.
If the CI lies strictly above 0, the improvement is considered statistically significant.

\paragraph{Results.}
Table~\ref{tab:bootstrap_scierc} shows the paired bootstrap results on SciERC.
DEPTH improves micro-F1 by $+8.6\%$ over Qwen-RLHF, and the 95\% CI of $\Delta$ is $[+6.8, +10.4]$, which confirms a statistically significant improvement.

\begin{table}[t]
    \centering
        \begin{subtable}[t]{0.53\linewidth}
        \centering
        \begin{tabular}{lcc}
            \toprule
             & \textbf{InstructUIE} & \textbf{InstructUIE} \\
             & \textbf{Correct} & \textbf{Wrong} \\
            \midrule
            \textbf{DEPTH Correct} & 967 & 322 \\
            \textbf{DEPTH Wrong}   & 138 & 48 \\
            \midrule
            \multicolumn{3}{c}{Exact McNemar (two-sided): $p = 5.14 \times 10^{-18}$} \\
            \bottomrule
        \end{tabular}
        \caption{McNemar's test on SciERC \texttt{NO-RELATION} subset $D_2$.}
        \label{tab:mcnemar_scierc}
        \end{subtable}
    \hfil
        \begin{subtable}[t]{0.42\linewidth}
        \centering
        \begin{tabular}{lcc}
            \toprule
            \textbf{Model} & \textbf{micro-F1} & \textbf{Gain ($\Delta$)} \\
            \midrule
            Qwen-RLHF          & 76.3           & \multirow{2}{*}{+8.6} \\
            DEPTH              & 67.7           & \\
            \midrule
            \multicolumn{3}{c}{95\% CI for $\Delta$: [+6.8, +10.4]} \\
            \bottomrule
        \end{tabular}
        \caption{Paired bootstrap on SciERC full test set $D$.}
        \label{tab:bootstrap_scierc}
        \end{subtable}
    \caption{Statistical significance tests for DEPTH vs.\ InstructUIE and Qwen-RLHF on SciERC.}
    \label{tab:significance_scierc}
\end{table}

\end{document}